%% file: main.tex
\documentclass[msc,deptreport,datasci]{infthesis}
\pdfoutput=1
\usepackage{breakcites}
\usepackage[pdftex]{graphicx}
\usepackage{xcolor}
\usepackage{url}
\usepackage{amsmath}
\usepackage{amsfonts}
\usepackage{multirow}
\usepackage{array}
\usepackage{caption}
\usepackage{subfigure}
\usepackage{mathrsfs}
\usepackage[colorlinks,linkcolor=black,anchorcolor=black,citecolor=black]{hyperref}

\begin{document}
\begin{preliminary}

\title{Emergence of Numeric Concepts in Multi-Agent Autonomous Communication}

\author{Shangmin Guo}

\include{abstract}

\maketitle

\include{acknowledgement}

\tableofcontents
\end{preliminary}

\include{chapter1}
\include{chapter2}
\include{chapter3}

\include{chapter4}
\include{chapter5}


\bibliographystyle{apalike}
\bibliography{main}

%
%
%

\end{document}

%% file: abstract.tex
\abstract{
    Natural language understanding is a long-standing topic in artificial intelligence. With the rapid development of deep learning, most of current state-of-the-art techniques in natural langauge processing are based on deep learning models trained with large-scaled static textual corpora. However, we human beings learn and understand in a different way. Thus, grounded language learning argues that models need to learn and understand language by the experience and perceptions obtained by interacting with enviroments, like how humans do.

    With the help of deep reinforcement learning techniques, there are already lots of works focusing on facilitating the emergence of communication protocols that have compositionalities like natural languages among computational agents population. Unlike these works, we, on the other hand, focus on the numeric concepts which correspond to abstractions in cognition and function words in natural language. 

    Based on a specifically designed language game, we verify that computational agents are capable of transmitting numeric concepts during autonomous communication, and the emergent communication protocols can reflect the underlying structure of meaning space. Although their encodeing method is not compositional like natural languages from a perspective of human beings, the emergent languages can be generalised to unseen inputs and, more importantly, are easier for models to learn. Besides, iterated learning can help further improving the compositionality of the emergent languages, under the measurement of topological similarity. Furthermore, we experiment another representation method, i.e. directly encode numerals into concatenations of one-hot vectors, and find that the emergent languages would become compositional like human natural languages.  Thus, we argue that there are 2 important factors for the emergence of compositional languages: i) input feature representations are inherently disentangled; ii) effective methods to amplify compositional inductive bias, e.g. iterated learning.
}

%% file: acknowledgement.tex
\section*{Acknowledgements}
Throughout the writing of this dissertation, I have received a great deal of support and assistance from many people. 

First of all, I would like to express my deepest appreciation to my supervisors, Dr. Ivan Titov and Prof. Kenny Smith, for their support for the proposal of this project and also their patient guidance, encouragement and advices. I am extremely lucky to have supervisors who cared so much about this work and proposed so much insights about the research topic.

I would also like to extend my deepest gratitude to my tutors, Mr. Serhii Havrylov and Dr. Stella Frank, whose expertise were invaluable in the formulating of the research topic and methodologies in particular. Without their insights and sagacities, completing this work would be much more difficult.

In particular, I greatly appreciate the support received through the collaborative work undertaken with Joshua Ren for the iterated learning part in this work as well as his insights for discussions about explaining the phenomena in experiments.

I am deeply indebted to all my friends in Edinburgh who opened their homes to me during my time at University of Edinburgh and who were always so helpful in numerous ways. Special thanks to Yan Yang, Yiyun Jin, Ruochun Jin, Muyang Liu, Wenbin Hu, Yuanhao Li and Jie Zhou.

I am extremely grateful to my beloved girlfriend, Siting Lu. Thank you for supporting me for everything and especially I cannot thank you enough for being with me throughout this experience. Thank you for all your companionship and care to me.

I cannot begin to express my heartfelt thank you to my Mum and Dad for always believing in me and encouraging me to follow my dreams as well as for helping in whatever way they could during this challenging period.

%% file: chapter1.tex
\chapter{Introduction}
\label{ch1:intro}

\begin{figure}[!h]
  \centering
  \includegraphics[width=0.8\textwidth]{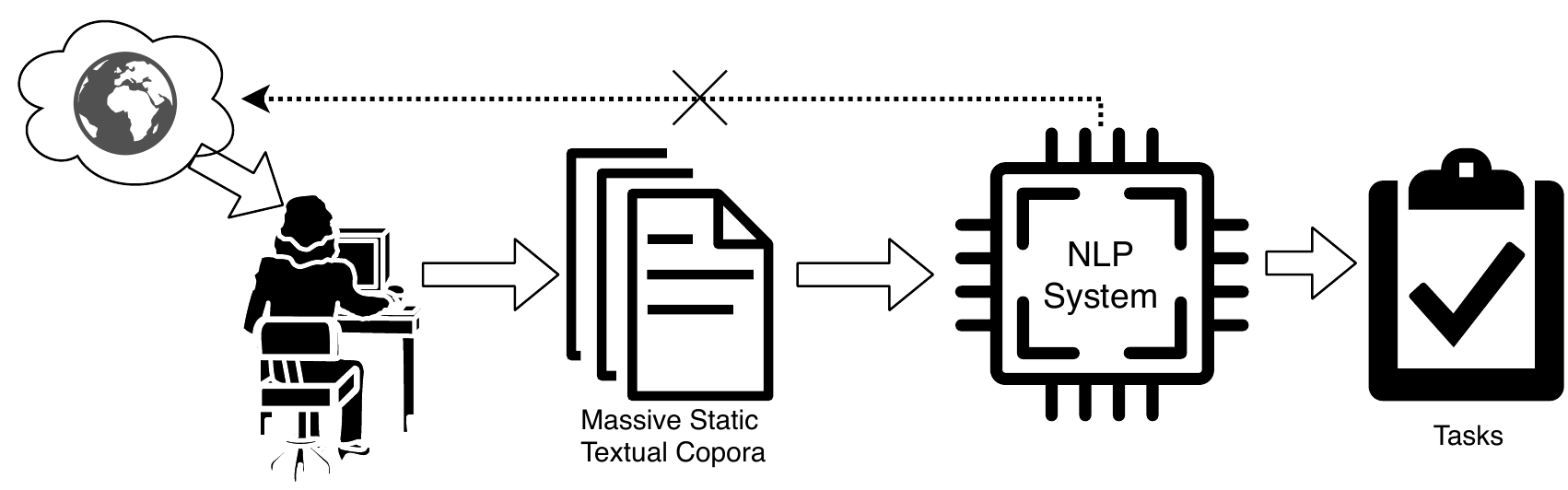}
  \caption{An overview of information flow in current NLP systems.}
  \label{fig1:nlpdiagram}
\end{figure}

Natural language processing (NLP) is an important and long-standing topic in artificial intelligence (AI), in which a core question is natural language understanding (NLU). With the rapid development of deep learning (DL), most current state-of-the-art methods in NLP, e.g. \cite{socher2013recursive, word2vec2013, kim2014cnn}, are based on DL models trained on massive static textual corpora. From an information processing perspective, the information flow of NLP-based human-computer interaction systems is illustrated in Figure \ref{fig1:nlpdiagram} given as follow. As the diagram shows, the input of NLP systems are various kinds of textual materials generated by human beings to describe their experiences and perceptions. Under such a perspective, symbols in natural languages are actually feature representations of the original experiences and perceptions, whereas most current NLP systems directly take these symbols as original features.

Considering the missing original experiences and perceptions, grounded language learning (GLL) argues that models need a grounded environment to learn and understand language\cite{matuszek2018grounded}. However, natural languages of the time have been developed for at least tens of thousands of years\cite{berwick2016only} and already became very sophisticated. Thus, to verify that computational agents can truly ground symbols in natural languages to corresponding experiences and perceptions, as well as be able to complete the specified tasks, it is necessary to help them to discover and develop various kinds of characteristics of natural language during autonomous communication of agents. There are already lots of works, e.g. \cite{batali1998computational, christiansen2003language, smith2003complex, hill2017understanding, havrylov2017emergence, yu2018interactive, kottur2017natural}, aiming to facilitate the emergence of ``natural language'' in multi-agent autonomous communication systems. However, one significant limitation of previous works is that, only referential objects/attributes in environments, e.g. shapes and colours, were considered and to which discrete symbols were grounded.

This project, on the other hand, aims to explore and analyse the grounding of abstractions which are non-referential\footnote{By non-referential, we mean that ``there is no concrete entity in the world (real or virtual) can be referred as".} in the original experiences and perceptions of human beings. However, as it is too huge a topic to tackle, our project is limited to numeric concepts which correspond to cardinal numerals in natural languages for the following reasons: i) numeral systems are relatively simple and self-contained\cite{james1999numeral}; ii) concepts related to cardinal numerals are more straightforward to model with numeric representations; iii) functions of emergent cardinal numerals can be formalised and verified more reliably in simulation.

In this work, our main contributions are given as follows:

\begin{enumerate}
  \item We propose a language game in which we can verify whether computational agents can communicate numeric concepts with each other, and successfully train agents to ``invent'' communication protocols that can autonomously transmit numeric concepts.
  \item We further analyse and discuss the structure of the emergent communication protocols, and improve the compositionality by transforming iterated leaning proposed by \cite{smith2003iterated} to train the DL models.
  \item We compare learning speeds of various kinds of languages as well as different representations, and propose an alternative hypothesis for explaining the emergence of words with different types and functions.
\end{enumerate}

All our codes are released at \href{https://github.com/Shawn-Guo-CN/EmergentNumerals}{https://github.com/Shawn-Guo-CN/EmergentNumerals}. 

%% file: chapter2.tex
\chapter{Background}
\label{ch2:background}

There are 2 almost disjointly developed research topics that motivates this project, i.e. computer simulation methods in evolutionary linguistics and multi-agent games in grounded language learning. Thus, in the following 2 sections, we will introduce works which are highly related to our project from these 2 different areas.

\section{Computer Simulation Methods in Evolutionary Linguistics}
\label{sec2.1:evolang}

One important issue in the field of evolutionary linguistics is to use quantitative methods to overcome the time limit on unpreserved pre-historic linguistic behaviours \cite{lieberman2006toward, evans2009myth}. Since it was first introduced by \cite{hurford1989biological}, computer simulation method has attracted a rapidly growing attention, e.g. \cite{hurford1998approaches, knight2000evolutionary, briscoe2002book, christiansen2003language, bickerton2009biological, cangelosi2012simulating}. As introduced in Chapter \ref{ch1:intro}, one of our objectives is to facilitate the discovery and development of various kinds of natural language phenomena of computational agents, which shares a same objective and motivation of computer simulation methods in evolutionary linguistics.

To imply and verify a linguistic theory, there are 2 necessary component: i) environments, in which agents can execute actions and communicate with each other; ii) pre-defined elementary linguistic knowledge that can be manipulated and altered by agents. Further, we could categorise the environments into the following 3 different types according to their simulation objectives:
\begin{itemize}
  \item \textit{Iterated learning} introduced by \cite{kirby1999function} which aims at simulating cultural transitions from generation to generation.
  \item \textit{Language games} introduced by \cite{wittgenstein2009philosophical} which takes the emergent communication protocols in cooperation between individuals as a prototype of language.
  \item \textit{Genetic evolution} introduced by \cite{briscoe1998language} which aims at simulating evolution of languages as a kind of natural selection procedure\cite{darwin1859origin}.
\end{itemize}

With environments and pre-defined elementary linguistic knowledge, computational agents can then learn bi-directional meaning–utterance mapping functions\cite{gong2013computer}. With different kinds of resulting linguistic phenomena, this simulation procedure can be broadly categorised into 2 classes:
\begin{itemize}
  \item lexical models, e.g. \cite{steels2005emergence, baronchelli2006sharp, puglisi2008cultural}, whose main concern is whether a common lexicon can form during the communication in agent community;
  \item syntactic and grammatical models, e.g. \cite{kirby1999function, vogt2005acquisition}, in which agents mainly aim to map meanings (represented in various ways) to utterances (either structured or unstructured ).
\end{itemize}

However, no matter how these mapping functions are learnt, e.g. by neural network models \cite{munroe2002learning} or by mathematical equations \cite{minett2008modelling, ke2008language}, the most basic elements of linguistics, e.g. meanings to communicate about and a signalling channel to employ, are all pre-defined.

In contract, although we also follow the framework of language games and train agents in an iterated learning fashion, the basic linguistic elements in our project are not provided from the outset any more and computational agents can specify meanings of symbols/utterances by themselves.

\section{Multi-agent Games in Grounded Language Learning}
\label{sec2.2:gll}

Unlike how we human beings learn and understand language, the current DL-based NLP techniques learn semantics from only large-scaled static textual materials. Thus, grounded language learning argues that computational agents also need to learn and understand languages by interacting with environments and grounding language into their experience and perceptions. With the recent rapid development of deep reinforcement learning (DRL), it has been shown that computational agents can master a variety of complex cognitive activities, e.g. \cite{mnih2015human, silver2017mastering}. Thus, several papers in grounded language learning apply DRL techniques to allow agents to learn or invent natural languages\footnote{Strictly speaking, ``invent natural language'' should be called as ``invent communication protocols sharing compositionality like natural languages''. However, as our project is to facilitate compositionality in multi-agent communication protocols, we thus call these emergent communication protocols a kind of ``language" invented by agents}, such as \cite{hermann2017grounded, mordatch2018emergence, havrylov2017emergence, hill2017understanding}.

To verify language abilities of computational agents, previous works in grounded language learning usually follow the framework of language games, of which are mainly variants of referential games introduced by \cite{lewis2008convention}, e.g. \cite{hermann2017grounded, havrylov2017emergence}. Also, some works are more motivated by the competition instead of cooperation such as \cite{cao2018emergent}.

From another perspective, based on the number of participated agents, we can broadly categorise language games in GLL into the following 2 types:

\begin{itemize}
  \item \textit{Single-agent games} usually need to be done by one agent and a human participator, in which the main concern is to explore how computational agents could learn the compositionality of semantics.
  \item \textit{Multi-agent games} are usually completed by an agent population, in which the main concern is to explore how various kinds of natural language phenomena emerge and evolve during communicating among agents.
\end{itemize}

However, like we mentioned in Chapter \ref{ch1:intro}, whichever kind of language game they follow in previous works of grounded language learning, the objects/attributes the symbols grounded to are all referential. We, on the other hand, aim to explore the feasibility of grounding symbols to non-referential objects (specifically, numeric concepts) during the game.

%% file: chapter3.tex
\chapter{Game, Models and Metrics}
\label{ch3:game_model}

In this chapter, we first describe the proposed language game and the definition of numerals in our game. We then introduce the architecture of models we trained and also the transformed iterated learning framework for training models.

\section{Game Description}
\label{sec3.1:game_description}

Unlike traditional simulation methods in evolutionary linguistics introduced in Section \ref{sec2.1:evolang}, there are 3 necessary components in our architecture and they are given as follows:

\begin{itemize}
  \item \textit{Environment}: To imply our linguistic assumption as well as make the size of environment limited and thus analysable, all perceptions in the established environment are sequences of objects represented by one-hot vectors. For ease of demonstration, we denote these objects as $o \in \mathcal{O}$ where $\mathcal{O} = \{A, B, C, \dots\}$ is the universal set of all kinds of objects in the following sections.
  \item \textit{Agents}: There are 2 kinds of agents in our project: i) \textit{speakers} $S$ that can observe objects in the environment and emit messages $m_i$; ii) \textit{listeners} $L$ that can receive the messages and generate a sequence of objects.
  \item \textit{Dynamics}: In this project, the dynamics mean not only the manually designed reward function for agents but also all elements related to training them, e.g. iterated learning and blank vocabulary. The details will be introduced in Subsection \ref{ssec3.2.3:loss_learning} and Subsection \ref{ssec3.2.4:iterated_learning}.
\end{itemize}

It worth mentioning that one premise of our project is that we do not have any assumption about the architecture of computational agents, and we focus more on the representations from environments as well as how agents are trained.

\subsection{Game Procedure}
\label{ssec3.1.1:game_procedure}

\begin{figure}[!h]
  \centering
  \includegraphics[width=0.8\textwidth]{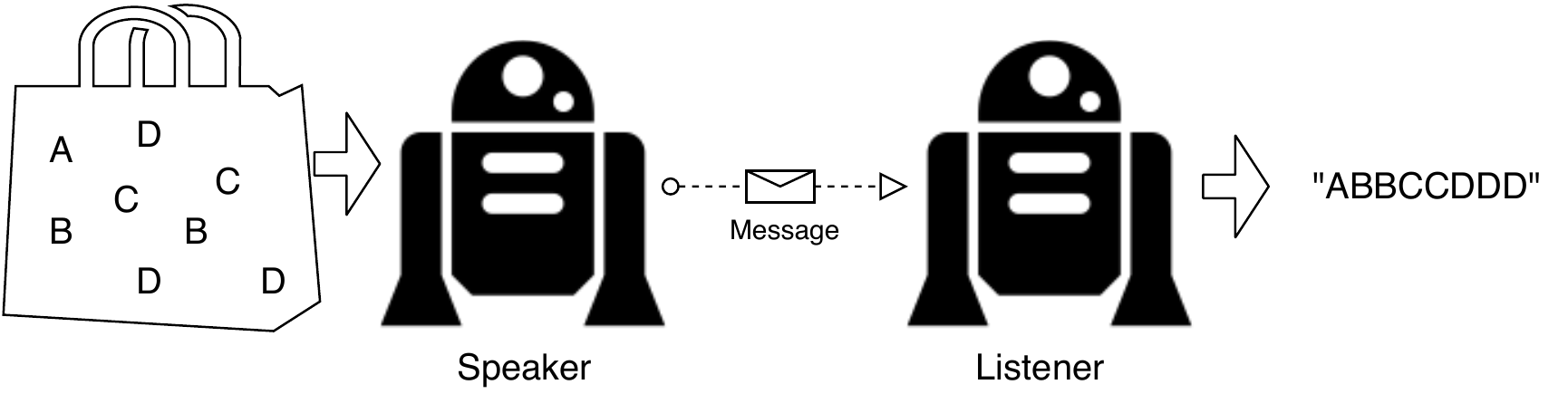}
  \caption{Diagram of Game Playing Procedure.}
  \label{fig2:game_procedure}
\end{figure}

 The first proposed game is to let listeners reconstruct sets of objects based on the messages transmitted by speakers, thus we call it ``Set-Reconstruct'' game. The overall view of the proposed Set-Reconstruct game is illustrated in Figure \ref{fig2:game_procedure} given as follow.

According to the steps of playing games at iteration $i$, the components of our games are illustrated as follows:
\begin{enumerate}
  \item Perceptions: the perception from environments is a \textbf{set} of objects, i.e. $s_i=\{o_{i_1}, o_{i_2}, \dots\ o_{i_n}\} \in \mathcal{S}$ where $n$ is the number of elements and $\mathcal{S}$ is meaning space.
  \item Speaker observation and message generation: after observing and encoding the perception, speaker $S$ would generate a message $m_i=\{t_{i_1}, t_{i_2}, \dots, t_{i_{|M|}}\} \in \mathcal{M}$ where $|M|$ is the maximum length of messages, $t_k \in V$ ($k \in {1, \dots, |V|}$) are selected from a randomly initialised vocabulary such that the symbols in the initially meaningless vocabulary whose size is $|V|$, and $\mathcal{M}$ is message space;
  \item Listener receiving message and perception reconstruction: after receiving and encoding the message $m_i$, the listener would generate a \textbf{sequence} $\hat{s}_i = \{\hat{o}_{i_1}, \hat{o}_{i_2}, \dots\ \hat{o}_{i_n}\}$ whose symbols are identical to those in the original perception $s_i$;
  \item Reward and parameter update: by comparing $s_i$ and $\hat{s}_i$, we take the cross-entropy between them as the reward for both listener and speaker and update parameters of both speaker and listener with respect to it.\footnote{Different ways of updating parameters are introduced in Section \ref{sec3.2:models}.}
\end{enumerate}

One thing that needs to be highlighted is that the perceptions $s_i$ are sets and thus order of objects would not make any difference. Further, we argue that the only important feature that need to be transmitted is actually the numbers of different objects which corresponds to the function of numerals in natural language.

\subsection{Functions of Numerals in the Game}
\label{ssec3.1.2:numeral_in_game}

Broadly speaking, numerals are words that can describe the numerical quantities and usually act as determiners to specify the quantities of nouns, e.g. "two dogs" and "three people". Also, under most scenarios, numerals correspond to non-referential concepts\cite{da2016wow}. Considering the objective of listeners $L$ in our language game, we define a numeral as a symbol $t^n$ at \textbf{position} $i$ indicating a function that reconstructs some object $o_i$ exactly $n$ times:

\begin{equation}
  t^n: o_i \rightarrow \{\overbrace{o_i, \dots, o_i}^{n \mbox{ elements}}\}
  \label{eq:3.1numeral_define}
\end{equation}

Note that, the meaning of a symbol is not only decided by itself but also its position in message, as $L$ would encode meanings of symbols according to their appearance in messages.

From the side of speakers $S$, a numeral is defined as a symbol $t^n$ at \textbf{position} $i$ that represents the numbers of specific object $o_i$, as we cannot tell whether agents realise the meanings of symbols are not related to their positions in the messages without specifically designed model architecture. Thus, we expect speaker $S$ would first learn to count the number of different objects and then encode them into a sequence of discrete symbols. As \cite{Siegelmann1992NN} shows that Recurrent Neural Networks (RNNs) are Turing-complete and Long-short Term Memory (LSTM) model proposed by \cite{hochreiter1997long} is a super set of RNN, it is safe to claim that LSTM is also Turing-complete and thus capable of counting numbers of objects.

\subsection{A Variant: Set-Select Game}
\label{ssec:3.1.3:refer_game}

\begin{figure}[!h]
  \centering
  \includegraphics[width=0.8\textwidth]{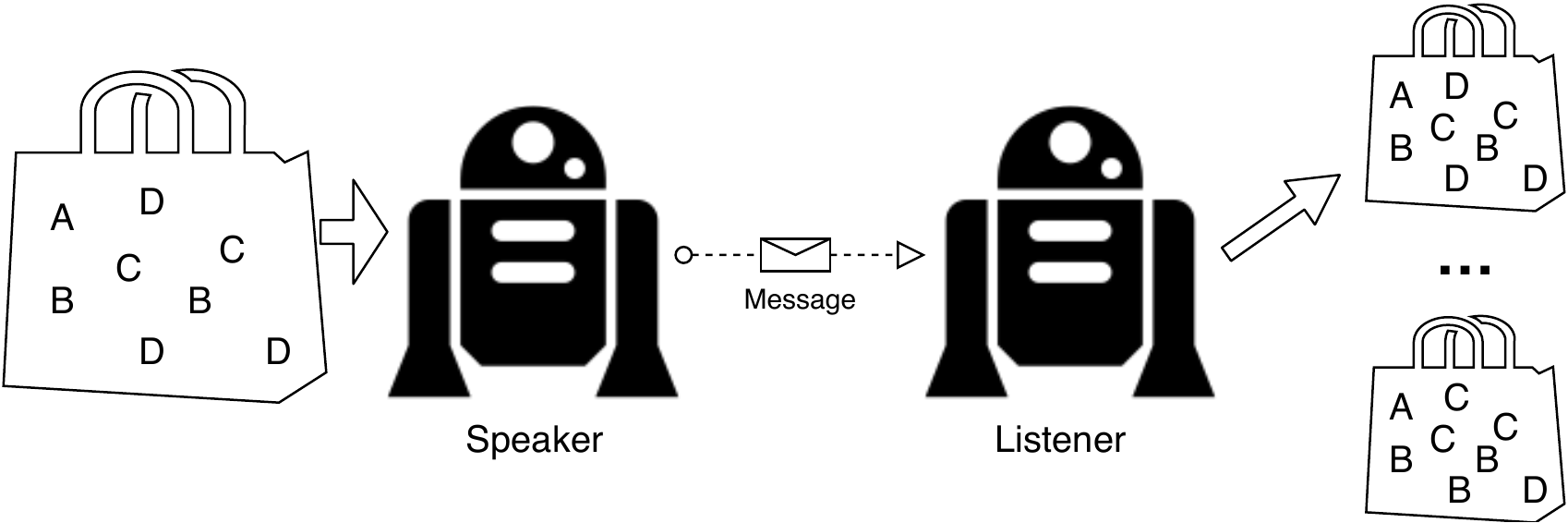}
  \caption{Diagram of Referential Game Playing Procedure.}
  \label{fig3:refer_game_procedure}
\end{figure}

We illustrate the Set-Select game, a referential variant of Set-Reconstruct game, in Figure \ref{fig3:refer_game_procedure} given above. The only difference is that listeners need to select the correct set of objects among several distractors \footnote{A distractor is a set that contains different numbers of objects as the correct one.} instead of reconstructing it.

\section{Proposed Models}
\label{sec3.2:models}

\begin{figure}[!h]
  \centering
  \includegraphics[width=0.9\textwidth]{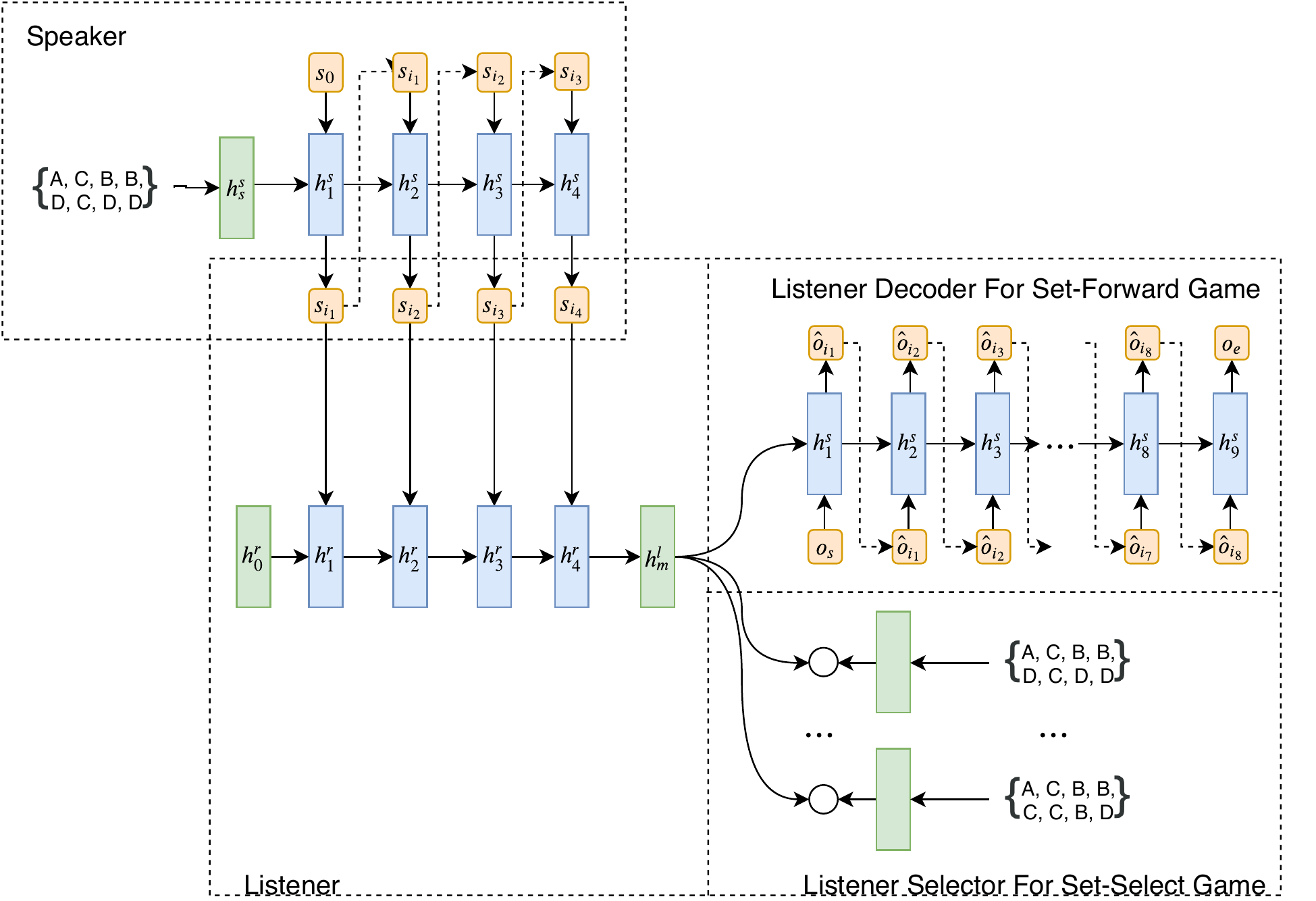}
  \caption{Overall Diagram of Model Architectures for Playing Games.}
  \label{fig4:model_arch}
\end{figure}

We illustrate the overall architecture of our models in Figure \ref{fig4:model_arch} given above.

A speaker $S$ consists of 2 components: i) a set encoder that takes a set of objects as input and outputs its vector representation $h_s^s$; ii) a standard LSTM sequence decoder that can generate a message $s_{i_1}, s_{i_2}, s_{i_3}, \dots$ based on $h_s^s$.

As for a listeners $L$, it would first encode messages with a LSTM sequence encoder and get the feature vector $h^l_m$. Then, in the Set-Reconstruct game, $L$ would take $h^l_m$ as the initial hidden state and predict a sequence of objects with a LSTM sequence decoder, which is shown by the right upper part of Figure \ref{fig4:model_arch}. As for in Set-Select game, $L$ would compare $h^l_m$ with several sets which are encoded by set encoders of $L$, and select the one shown to $S$ based on the dot product between $h^l_m$ and feature vectors of all candidate sets.

Further details are shown in the following subsections.

\subsection{Speaker}
\label{ssec3.2.1:speaker}

The architecture of our speaking agents is very similar to the Seq-to-Seq model proposed by \cite{sutskever2014sequence} except that replace the encoder for input sequences with a set encoder whose details are introduced in the following subsubsection. As Seq-to-Seq model is quite popular nowadays, we skip details about how to generate sequences which correspond to the messages in our games, and focus on how to encode sets of objects.

\subsubsection{Set Encoder}
\label{sssec3.2.1.1:set_encoder}

Our set encoder shares an almost same architecture of inputting sets proposed by \cite{vinyals2015order}. However, as there is an addition in $softmax$ function and it would introduce counting bias into the feature representation of sets, we replace equation (5) in \cite{vinyals2015order} with the following operation in order to avoid exposing counting system to models:

\begin{equation}
  a_{i,t} = \sigma(e_{i,t})
  \label{eq3.2.1.1:sigmoid_to_replace_softmax}
\end{equation}
where $\sigma$ is sigmoid function.

Thus, assume the input for speaker $S$ is a set $s_i=\{o_{i_1}, o_{i_2}, \dots\ o_{i_n}\}$. The first step is to read the set $s_i$ as a sequence and project all objects to dense vectors by an embedding layer. Based on the sequence $\{w_{i_1}^s, w_{i_2}^s, \dots\ w_{i_n}^s\}$ (where $w_{i_k}^s$ is the embedding vector of $o_{i_k}$ for speaker where $k\in \{1, \dots, n\}$), the calculation of $h_s^s$ can be given as follows:

\begin{equation}
  \begin{split}
    e_{i, t}^s & = f(q_{t-1}^s, w_i^s) \\
    a_{i, t}^s & = \sigma(e_{i,t}^s) \\
    r_t^s & = \sum_i a_{i,t}^s w_i^s  \\
    q_t^s &= LSTM(r_t, q_{t-1}^s, c_{t-1}^s)
  \end{split}
  \label{eq3.2.1.2:speaker_hidden_calculation}
\end{equation}
where $t\in \{1, \dots, T\}$ is the number of attention times, $f$ is an affine layer, $q^s_{t}$ and $c^s_t$ are hidden and cell states in LSTM respectively.

Besides, in our implementation, $T$ is set to be the same as the number of all types of objects, as we want to help models to represent number of each kind of objects as features in the vector representation of input set.

\subsubsection{Message Generator}
\label{sssec3.2.1.2:msg_generator}

To generate the message $m_i$, we follow \cite{havrylov2017emergence} and adopt a LSTM-based sequence decoder with 2 different kinds of sampling mechanisms: i) direct sampling that directly sample from the corresponding categorical dist6ribution specified by $softmax(Wh_k^s + b)\ \forall k\in {1, 2, \dots, |M|}$; ii) Gumbel-softmax estimator proposed by \cite{jang2016categorical} with straight-through trick introduced by \cite{bengio2013estimating}. Beside, the learning mechanisms also vary for these 2 different sampling methods, which is further discussed in Subsection \ref{ssec3.2.3:loss_learning}.

Note that the length of each message $m_i$ is fixed to $|M|$ and symbols $t_{i_1},\dots,t_{i_|M|}$ are all one-hot vectors that represent different discrete symbols. The effect of number of all discrete message symbols $|V|$ and length of messages $|M|$ on the emergent language is further discussed in Chapter \ref{ch4:results_analysis}.

\subsection{Listener}
\label{ssec3.2.2:listeners}

The architectures of listening agents are specifically designed for handling different kinds of tasks/games and thus vary from Set-Reconstruct game to Set-Select game.

\noindent\textbf{Listener in Set-Reconstruct Game}: The listener in Set-Reconstruct game has exactly the same architecture as Seq-to-Seq model proposed by \cite{sutskever2014sequence}. And, when combined with speaker model, the overall model is called ``Set2Seq2Seq''.

\noindent\textbf{Listener in Set-Select Game}: The listener in Set-Select game would also first encode messages with a LSTM like it is in standard Seq-to-Seq model. However, as it needs to select among several candidates, it also needs to encode all these sets with Set Encoder introduced in Subsection \ref{sssec3.2.1.1:set_encoder}. Then, the listener would make predictions based on the dot-products between embedding of message $h^r_m$ and embeddings of each set of objects. Similarly, when combined with speaker model, the overall model is called as ``Set2Seq2Choice''.

\subsection{Loss/Reward and Learning}
\label{ssec3.2.3:loss_learning}

\textbf{In Set-Reconstruct game}, as the predictions of listeners are a sequence of objects $\hat{s}_i=\{\hat{o}_{i_1}, \dots, \hat{o}_{i_n}\}$, we use cross-entropy between the original set and the predicted sequence as the objective function that needs to be minimised. Formally,

\begin{equation}
  \mathcal{L}_{\theta^S, \theta^L}(o_{i_1}, \dots, o_{i_n}) =\mathbb{E}_{m_i\sim p_{\theta^S}(\cdot|s_i)} \left[ -\sum_{k=1}^{n} o_{i_k} \log(p(\hat{o}_{i_k}|m_i, \hat{o}_{-i_k})) \right]
  \label{eq3.2.3.1:cross_entropy_seq}
\end{equation}
where $\hat{o}_{-i_k}$ represent all predicted objects preceding $\hat{o}_{i_k}$.

\noindent\textbf{In Set-Select game}, we still use the cross entropy between the correct candidate and  as the loss to minimise, i.e.

\begin{equation}
  \mathcal{L}_{\theta^S, \theta^L}(s_i) = \mathbb{E}_{m_i\sim p_{\theta^S}(\cdot|s_i)} \left[-\sum_{k=1}^{C} s_i log(p(c_k)) \right]
  \label{eq3.2.3.2:cross_entropy_choose}
\end{equation}
where $c_k$ is the predicted logit score for candidate $k$ among $C$ candidates.

In the case that we use Gumbel-softmax to approximate sampling messages from speaker $S$, parameters $\theta^S$ and $\theta^L$ are learnt by back-propagation. In the case that we use direct sampling, $\theta^L$ is still learnt by back-propagation, where as $\theta^S$ is learnt by REINFORCE estimator \cite{williams1992simple} with cross-entropy scores as rewards.

\subsection{Neural Iterated Learning}
\label{ssec3.2.4:iterated_learning}

The evolutionary linguistic community has already studied the origins and metrics of language compositionality since \cite{kirby2002emergence} which points out a cultural evolutionary account of the origins of compositionality and proposes iterated learning to model this procedure. Thus, to facilitate the emergence of compositionality among the autonomous communication between agents, we also trained our agents in a iterated learning fashion. In the original iterated learning, an agent can both speak and listen. However, in this project, agent can be either a speaker or a listener, not both at the same time. Thus, we slightly transform the iterated learning framework and call the following one ``neural iterated learning'' (NIL).

Following the overall architecture of iterated learning, we also train agents generation by generation. In the beginning of each generation $t$, we would randomly re-instantiate a new speaker $S_t$ and a new listener $L_t$ and then execute the following 3 phases:

\begin{enumerate}
  \item \textbf{Speaker Learning phase}: During this phase, we would train $S_t$ with the set-message pairs generated by $S_{t-1}$, and the number of epochs is set to be fixed. Note that there is no such phase in the initial generation, as there is no set-message pair for training $S_t$.
  \item \textbf{Game Playing phase}: During this phase, we would let $S_t$ and $L_t$ cooperate to complete the game and update $\theta^S_t$ and $\theta^L_t$ with loss/reward illustrated in previous section, and use early-stopping to avoid overfitting.
  \item \textbf{Knowledge Generation phase}: During this phase, we would feed all $s_i$ in training set into $S_t$ and get corresponding messages $m_i$. Then, we would keep the sampled ``language'' for $S_{t+1}$ to learn.
\end{enumerate}

\subsection{Baseline Models}
\label{ssec3.2.5:baselines}

To get the upper bounds of our multi-agent communication systems, we remove the communication between speaker and listener to be the baseline models.

In Set-Reconstruct game, our baseline is Set-to-Seq model which first encodes the input set $s_i$ with the set encoder introduced in subsection \ref{sssec3.2.1.1:set_encoder} and then directly generate the predicted sequence $\hat{s}_i$ following the sequence generation in standard seq-to-seq model.

As for in Set-Select game, our baseline is Set-to-Choose model, in which speaker directly transmit representation vector $h^s_s$ of set $s_i$ to listener. And, listener compare $h^s_s$ with all candidate sets to make a selection.

\section{Compositionality and Metrics}
\label{sec3.3:metrics}

With the recent rapid development of grounded language learning, measuring the compositionality of emergent communication protocol attracts more and more attention nowadays, e.g. \cite{andreas2019measuring}, \cite{lowe2019pitfalls}.

First of all, to better define compositionality, we argue that if a language is said to be perfect compositional, then it should satisfy the following 2 properties:

\begin{itemize}
    \item \textbf{Mutual Exclusivity}: Symbols describing different values of a same property should be mutually exclusive to each other. For example, ``green'' and ``red'' are both used to describe colour of an object and they should not appear at the same time as an object can not be green and red at the same time.
    \item \textbf{Orthogonality}: Appearance of symbols for describing a property should be independent from the appearance of symbols used to describe another property. For example, the appearance of symbols used for describing colours of objects should be independent from the appearance of symbols used for describing shapes of objects.
\end{itemize}

As the setting of our game is simple and the space size is limited, we follow \cite{brighton2006understanding} and take the topological similarity between meaning space (space of all sets of objects) and message space as the metric of compositionality. Briefly speaking, as much of language is neighbourhood related, i.e. nearby meanings tend to be mapped to nearby messages, the compositionality of language can be measured as the correlation degree between distances of meanings and distances of corresponding messages. For example, the meaning of set $\{A,A,A,B,B\}$ is closer to $\{A,A,B,B\}$ than $\{A,A,A,A,B,B,B\}$. In natural language (which is perfectly compositional), messages for  $\{A,A,A,B,B\}, \{A,A,B,B\}, \{A,A,A,A,B,B,B\}$ are ``3A2B'', ``2A2B'' and ``4A3B''\footnote{In Chapter \ref{ch4:results_analysis}, we would illustrate messages with lower case alphabets. To make them easier to understand, we use natural language here.} respectively. However, in a non-compositional language, the messages may be ``5B5A'', ``1C2E'' and ``3A4C'',  which are randomly sampled mappings between meaning space and message space.

In order to calculate the topological compositionality, we need define the distance metric for meaning space and message space respectively. Thus, for an input set $s_i$, we could first count the number of each kind of object and then concatenate the Arabic numerals as the meaning sequence. Take a set $s_i=\{A, A, A, B, B\}$ for example, the corresponding meaning sequence would be ``32'' as there are 3 $A$ and 2 $B$ in $s_i$.\footnote{Again, the appearing order of objects would not effect the meaning sequence of a set.} As for the message space, we have several different settings which are further illustrated in subsection \ref{ssec4.2.2:topo_sim}, and edit distance as in \cite{brighton2006understanding} is also included.

Meanwhile, as we could perfectly encode the meaning of a set into natural language, we could take the speaker as a machine translation model that translates a meaning represented in natural language into emergent language invented by computational agents themselves. Inspired by this point of view, we could also use BLEU score proposed by \cite{papineni2002bleu} as a metric of semantic similarities between messages. For the sets that share more similar meanings, we expect their corresponding messages to share more uni-grams or bi-grams or so on. Following the above example, in a perfectly compositional language, as $\{A,A,A,B,B\}$ locates very close to $\{A,A,B,B\}$, their messages (``3A2B'' and ``2A2B'') share $3$ uni-grams (``A'', ``B'' and ``2'') and $2$ bi-grams (``A2'' and ``2B'') in common. However, in a non-compositional language, e.g. in which the messages for $\{A,A,A,B,B\}$ and $\{A,A,B,B\}$ are ``5B5A'' and ``1C2E'' respectively, the messages share no uni-gram and bi-gram in common.

In our case, the BLEU score between $m_i$ and $m_j$ is calculated as follow:

\begin{equation}
  BLEU(m_i, m_j) = 1 - \sum_{n=1}^{N} \omega_n \cdot \frac{\mbox{Number of common } n\mbox{-grams}}{\mbox{Number of total different } n\mbox{-grams}}
  \label{eq3.3.1:bleu_score}
\end{equation}
where $n$ is the size of $n$-grams and $\omega_n$ is the weight for similarity based on $n$-grams. In the following discussions, we would denote BLEU score based on $n$-grams as BLEU-$n$, e.g. BLEU score based on uni-grams would be represented as BLEU-1.

%% file: chapter4.tex
\chapter{Experiment Results and Discussion}
\label{ch4:results_analysis}


\section{Emergence of Language without Iterated Learning}
\label{sec4.1:emergence}

First of all, we have to verify that the agents can successfully address the problems by communicating with discrete symbols. After tried several different settings, to avoid that the success of agents depends on fine-tuning hyperparameters, we find that it is better to make the size of message space much larger than the size of meaning space. Thus, we set the size of message space $|V|^{|M|}$ to be 100 times of the meaning space $|N_{o}|^{|\mathcal{O}|}$ and show the performance of both ``Set2Seq2Seq'' and ``Set2Seq2Choice'' in Table \ref{tab4.1:game_performance}. In the table, $|M|$ is the length of messages, $|V|$ is the size of vocabulary\footnote{Note that the meaning of ``vocabulary" is not like it is in traditional NLP, but refers to the set of initially meaningless symbols that can be used for communication.} for message, $|\mathcal{O}|$ is the number of all kinds of objects and $|N_o|$ is the maximum number of a single kind of object.

Additionally, as the training procedure is time-consuming, all the shown performance are based on a single run, and thus the effects from hyperparameters and randomness cannot be completely filtered out. However, as we did not intentionally fine-tune the hyperparameters and we focus on the emergent communication protocols, we believe that the variabilities of performance is limited and would not affect our following discussions.

\begin{table}[!h]
    \centering
    \begin{tabular}{|c|c|c|c|}
        \hline
        Model                           & Sampling Method & Performance & Game Setting      \\ \hline
        \multirow{3}{*}{Set2Seq2Seq}    & Gumbel          & 99.89\%     & \multirow{3}{1.5in}{$|M|=8$, $|V|=10$, $|\mathcal{O}|=6$, $|N_{o}|=10$} \\ \cline{2-3}
                                        & REINFORCE       & 89.89\%     &                   \\ \cline{2-3}
                                        & SCST            & 98.67\%     &                   \\ \hline
        \multirow{3}{*}{Set2Seq2Choice} & Gumbel          & 100\%       & \multirow{3}{1.5in}{$|M|=6$, $|V|=10$, $|\mathcal{O}|=4$, $|N_{o}|=10$} \\ \cline{2-3}
                                        & REINFORCE       & 76.45\%     &                   \\ \cline{2-3}
                                        & SCST            & 83.26\%     &                   \\ \hline
        \end{tabular}
    \caption{Performance of Models and Corresponding Game Settings.}
    \label{tab4.1:game_performance}
\end{table}

Beside the ``REINFORCE'' and ``Gumbel'' sampling methods introduced in subsection \ref{sssec3.2.1.2:msg_generator}, we also tried the self-critic sequence training proposed by \cite{rennie2017self} as a baseline for REINFORCE algorithm, which is denoted by ``SCST''. Briefly speaking, SCST utilises the output of its own test-time inference algorithm to normalize the rewards received instead of estimating a separate “baseline”.

Based on the performance shown in Table \ref{tab4.1:game_performance}, it is clear that Gumbel is the most stable sampling method on all different settings. Thus, unless specifically stated, the following experiments and discussions are all based on training agents with Gumbel-softmax as message sampling method.


\section{Structure of Emergent Language}
\label{sec4.2:structure_emergent_lan}

\subsection{Emergent Languages in Various Games}
\label{ssec4.2.1:emergent_languages}

After verifying that computational agents are able to complete games through communication, we are curious about the messages produced during their communication. However, unlike what was shown by the previous works in grounded language learning, e.g. \cite{hill2017understanding} and \cite{mordatch2018emergence}, the emergent language during both Set-Reconstruct and Set-Select games are not ``perfectly'' compositional, which will be illustrated later. From our perspective, one alternative explanation for this phenomenon is that $|M| > |\mathcal{O}|$ in our game settings, which makes proportion of holistic languages \footnote{A holistic language is a language that needs to be learned as a whole and should not be analysed or compartmentalized. In this work, holistic languages are generated by randomly sampling mappings between meaning space and message space.} much larger than the proportion of compositional languages \cite{brighton2002compositional}, and thus it becomes very hard to find compositional languages.

To have give an intuitive demonstration of the emergent language, we list all messages transmitted in a Set-Reconstruct game where $|\mathcal{O}|=2, |N_o|=5, |M|=4, |V|=10$ in Table \ref{tab4.2:emregent_language_generation} given as follow. In the table, the first row and first column are the basic elements of meanings and each cell is the corresponding message for that meaning. Take cell ``1A2B'' for example, the original input set is $s_i=\{A,B,B\}$ and the corresponding message $m_i$ is ``ttvz''. Note that the alphabets in the message do not correspond to any symbol in natural language.

\begin{table}[!h]
    \centering
    \begin{tabular}{|c|c|c|c|c|c|c|}
        \hline
           & 0A   & 1A   & 2A   & 3A   & 4A   & 5A   \\ \hline
        0B &      & txtt & txzt & xtzz & xzzz & xxvx \\ \hline
        1B & ttxt & ttxz & tzzz & ztzv & zzvz & vzxv \\ \hline
        2B & tttx & ttvz & tzhz & tvzv & zvhv & vvvz \\ \hline
        3B & tttv & ttvw & thzv & tvwv & hvzv & wvzv \\ \hline
        4B & ttht & thtw & thwz & hhvz & hwvv & wwvv \\ \hline
        5B & tthh & thhh & thww & hhwh & hwww & wwww \\ \hline
        \end{tabular}
    \caption{An emergent language in a Set-Reconstruct game.}
    \label{tab4.2:emregent_language_generation}
\end{table}

Based on the 2 properties of compositional languages illustrated in Section \ref{sec3.3:metrics}, we could see that the emergent language shown in Table \ref{tab4.2:emregent_language_generation} satisfies neither of mutual exclusivity nor orthogonality. To be specific, there is no common substrings of messages in every column/row, and some substrings, e.g. ``tt'' in column ``0A'', may be used in multiple columns/rows. Thus, the emergent language is not a perfectly compositional one as we expect. Thus, as we can see from Table \ref{tab4.2:emregent_language_generation}, there is no clear compositional structure in it.

However, as Set-Reconstruct game is a generation task, one possible hypothesis is that the agents may transmit more than numeric concepts in order that listeners could generate the original input. Thus, to verify whether this is the case, we illustrate an emergent language in a Set-Select game whose settings are exactly the same as the Set-Reconstruct game illustrated above, i.e. $|\mathcal{O}|=2, |N_o|=5, |M|=4, |V|=10$. The meanings and corresponding messages are shown in Table \ref{tab4.3:emregent_language_referential} given as follow.

\begin{table}[!h]
    \centering
    \begin{tabular}{|c|c|c|c|c|c|c|}
        \hline
           & 0A   & 1A   & 2A   & 3A   & 4A   & 5A   \\ \hline
        0B &      & xxxv & xxvy & xvyy & vyxy & vyyy \\ \hline
        1B & xxxx & xxzx & xwxv & xvvv & vvxx & vvyy \\ \hline
        2B & xxyx & xqxx & xzxz & xwwv & vwvx & vvvv \\ \hline
        3B & xyxy & xqyx & xqqx & zxzz & wwwx & wvwv \\ \hline
        4B & yxyx & yxqy & qxqy & qzxq & zzxz & zwwz \\ \hline
        5B & yyxy & yyqy & qyyy & qqyy & zqqq & zzzz \\ \hline
        \end{tabular}
    \caption{An emergent language in a Set-Select game.}
    \label{tab4.3:emregent_language_referential}
\end{table}

Based on the message contents in Table \ref{tab4.3:emregent_language_referential}, we could find that the referential game does not necessarily make the emergent language perfectly compositional.

According to \cite{kottur2017natural}, another alternative probability is that the message space is much larger in the previous game settings and thus it is over-complete for agents to encode the sets of objects in a compositional fashion. Thus, we re-train that agents with $|\mathcal{O}|=2, |N_o|=5, |M|=2, |V|=10$ \footnote{Here, the message space is still larger than meaning space, as we again do not want to make the success of agents depend on fine-tuning hyperparameters.} (where the size of meaning and message space are $35$ and $100$ respectively), and the emergent language is shown in Table \ref{tab4.4:emregent_language_referential2}. As we can see, the smaller message space does not necessarily facilitate the emergence of compositional language in Set-Select game.

\begin{table}[!h]
    \centering
    \begin{tabular}{|c|c|c|c|c|c|c|}
        \hline
           & 0A & 1A & 2A & 3A & 4A & 5A \\ \hline
        0B &    & zv & vz & vv & vv & xv \\ \hline
        1B & zy & zu & zw & wz & wv & xw \\ \hline
        2B & yz & uu & zz & zt & ww & wx \\ \hline
        3B & yz & uy & uq & qz & tz & tw \\ \hline
        4B & yx & yy & ur & qq & qt & tt \\ \hline
        5B & xy & yr & ry & rx & xq & xt \\ \hline
        \end{tabular}
    \caption{Another emergent language in a Set-Select game.}
    \label{tab4.4:emregent_language_referential2}
\end{table}

\subsection{Topological Similarities}
\label{ssec4.2.2:topo_sim}

As introduced in subsection \ref{sec3.3:metrics}, we measure the topological similarity between meaning space and message space as a measure of compositionality. We list compositionality scores under different kinds of metrics in Table \ref{tab4.4:topo_sim_lans} given as follow.

\begin{table}[!h]
    \centering
    \begin{tabular}{|c|c|c|c|c|}
        \hline
                          & Ham+Edit & Ham+BLEU & Euclid+Edit & Euclid+BLEU \\ \hline
        Compositional     & 1.00     & 0.61     & 0.38        & 0.24        \\ \hline
        Set-Reconstruct   & 0.32     & 0.27     & 0.60        & 0.65        \\ \hline
        Set-Select        & 0.13     & 0.16     & 0.45        & 0.52        \\ \hline
        Holistic          & -0.04    & -0.04    & 0.01        & 0.00        \\ \hline
    \end{tabular}
    \caption{Topological similarity scores of different languages.}
    \label{tab4.4:topo_sim_lans}
\end{table}

\noindent\textbf{Ham+Edit}: We first follow the distance metrics in \cite{brighton2006understanding}: i) use hamming distances between meaning sequences as the similarity metric for meaning space; ii) use edit distances between corresponding messages as the similarity metric for message space.

\noindent\textbf{Ham+BLEU}\footnote{Without special declaration, we use ``BLEU'' to represent weighted average between BLEU-1 and BLEU-2, i.e. $0.5\times \mbox{BLEU-1} + 0.5\times\mbox{BLEU-2}$.}: In this setting, we use: i) hamming for meaning space too; ii) BLEU score illustrated in Section \ref{sec3.3:metrics} as the the similarity metric for message space.

\noindent\textbf{Euclid+Edit}: In this setting, we use: i) Euclidean distance as the metric for meaning space, e.g. Euclidean distance between ``4A2B'' and ``1A3B'' is \\ $\sqrt{(4-1)^2 + (2-3)^2}=\sqrt{10}$; ii) edit distance for message space.

\noindent\textbf{Euclid+BLEU}: In this setting, we use: i) Euclidean distance for meaning space; ii) BLEU score illustrated in Section \ref{sec3.3:metrics} for message space.

To get the upper bound and lower bound of compositionality, we specifically designed: i) a perfectly compositional language, in which the message is exactly the same as meaning sequence, e.g. ``4A2B'' is represented as ``wsyr'' ($\mbox{A}\rightarrow s, \mbox{B} \rightarrow r, \mbox{4} \rightarrow w, \mbox{2} \rightarrow y$); ii) a holistic language, in which messages are randomly generated.

Then, from the above results, we could see that although the emergent languages in Set-Reconstruct and Set-Select games gain low topological similarity scores under Hamming distance for meaning space, they obtain much higher similarity scores under Euclidean distances for meaning space. As for the compositional language, they obtain very low scores under Euclidean distances, which is caused by that the numerals in natural language encode numeric concepts as different symbols and thus the edit/BLEU distance between messages are all the same for different meanings. For example, although meaning ``2A1B'' is closer to ``1A1B'' than meaning ``4A1B'', the edit distance between ``yszr'' (message for ``2A1B'') and ``zszr'' (message for ``1A1B'') is exactly the same as the edit distance between ``wszr'' (message for ``4A1B'') and ``zszr''.

\subsection{Significance Test of Same Numeric Concepts}
\label{ssec4.2.3:significance_test}

Although the compositionality of emergent language is not like our natural language, we could also see that it may reflect the underlying structure of meaning space. Thus, to further verify this point of view, we further verify whether messages for meaning pairs that share same numeric concepts are more similar. To do this, we established 2 different datasets: i) meaning pairs sharing exactly same numeric concepts, e.g. ``4A3B'' and ``3A4B'', and corresponding BLEU similarity scores for their messages; ii) pairs of meaning sequences that share no numeric concept, e.g. ``4A3B'' and ``5A1B'', and corresponding BLEU similarity for their messages. Thus, we have 1,190 ($2 \times (35 \times 34 \div 2)$) meaning pairs in total, of which half share same numeric concepts and the other half do not. We then calculate the BLEU-1/2/3 scores of the messages of these pairs and test whether these BLEU scores are correlated to sharing same numeric concepts.

Then, we establish the following hypotheses for significance test:

\begin{itemize}
    \item \textbf{Null hypothesis}: The BLEU scores between messages are independent from whether meaning pairs share same numeric concepts.
    \item \textbf{Alternative hypothesis}: The BLEU scores between messages are \textbf{not} independent from whether meaning pairs share same numeric concepts.
\end{itemize}

To test whether there is a correlation between BLEU scores and sharing same numeric concepts, we calculate the Spearman correlation coefficient and the corresponding $p$-values. The results got on different types of languages based on different $n$-grams are given in Table \ref{tab4.5:p-values} as follow. \footnote{It is clear that an effective language would not contain identical messages for different meanings. As the maximum length of messages is 4 and there is no identical messages in all languages, we thus skip the BLEU-4 score (which would 0 for all languages) here. }

\begin{table}[!h]
    \centering
    \begin{tabular}{|l|c|c|c|c|c|c|}
    \hline
    \multicolumn{1}{|c|}{\multirow{2}{*}{Language Type}} & \multicolumn{2}{c|}{BLEU-1}      & \multicolumn{2}{c|}{BLEU-2}                                      & \multicolumn{2}{c|}{BLEU-3}    \\ \cline{2-7}
    \multicolumn{1}{|c|}{}                               & $\rho$  & $p$-value              & $\rho$ & $p$-value                                               & $\rho$ & $p$-value             \\ \hline
    Compositional                                        & $0.96$  & $7.84 \times 10^{-39}$ & $0.28$ & $2.01\times 10^{-2}$                                   & $0.28$ & $2.01\times 10^{-2}$ \\ \hline
    Emergent-R                                           & $0.27$  & $2.42\times 10^{-2}$  & $0.26$ & $2.74\times 10^{-2}$                                   & $0.33$ & $5.07\times 10^{-3}$ \\ \hline
    Emergent-S                                           & $0.29$  & $1.49\times 10^{-2}$  & $0.38$ & $1.16\times 10^{-3}$                                   & $0.38$ & $1.06\times 10^{-3}$ \\ \hline
    Holistic                                             & $-0.08$ & $4.93\times 10^{-1}$  & $0.05$ & $6.77 \times 10^{-1}$ & $0.22$ & $6.10\times 10^{-2}$ \\ \hline
    \end{tabular}
    \caption{$\rho$ and $p$-values of different types of languages. $\rho$ represents the Spearman correlation coefficient between meanings and messages, and the ``Emergent-R'' and ``Emergent-S'' are emergent languages in Set-Reconstruct game and Set-Select game respectively. }
    \label{tab4.5:p-values}
\end{table}

The $p$-values under BLEU-1/2/3 for compositional language as well as emergent languages in both Set-Reconstruct and Set-Select games are all smaller than $0.05$. Thus, it is safe to reject null hypothesis and accept the alternative hypothesis. That is, The BLEU scores between messages are highly correlated with whether their meaning pairs share same numeric concepts. To be more precise, as we can tell from Table \ref{tab4.5:p-values}, the messages of meaning pairs sharing same numeric concepts have more uni-grams, bi-grams  and tri-grams in common.

\subsection{Generalisation of Emergent Language}
\label{ssec4.2.4:emergent_lan_generalise}

A further question about the structure of emergent languages is to verify whether the emergent language can be generalised to unseen sets. If the emergent languages can be generalised, we then could say that these languages do capture the structure of meaning spaces.

\begin{figure}[!h]
    \centering
    \includegraphics[width=\textwidth]{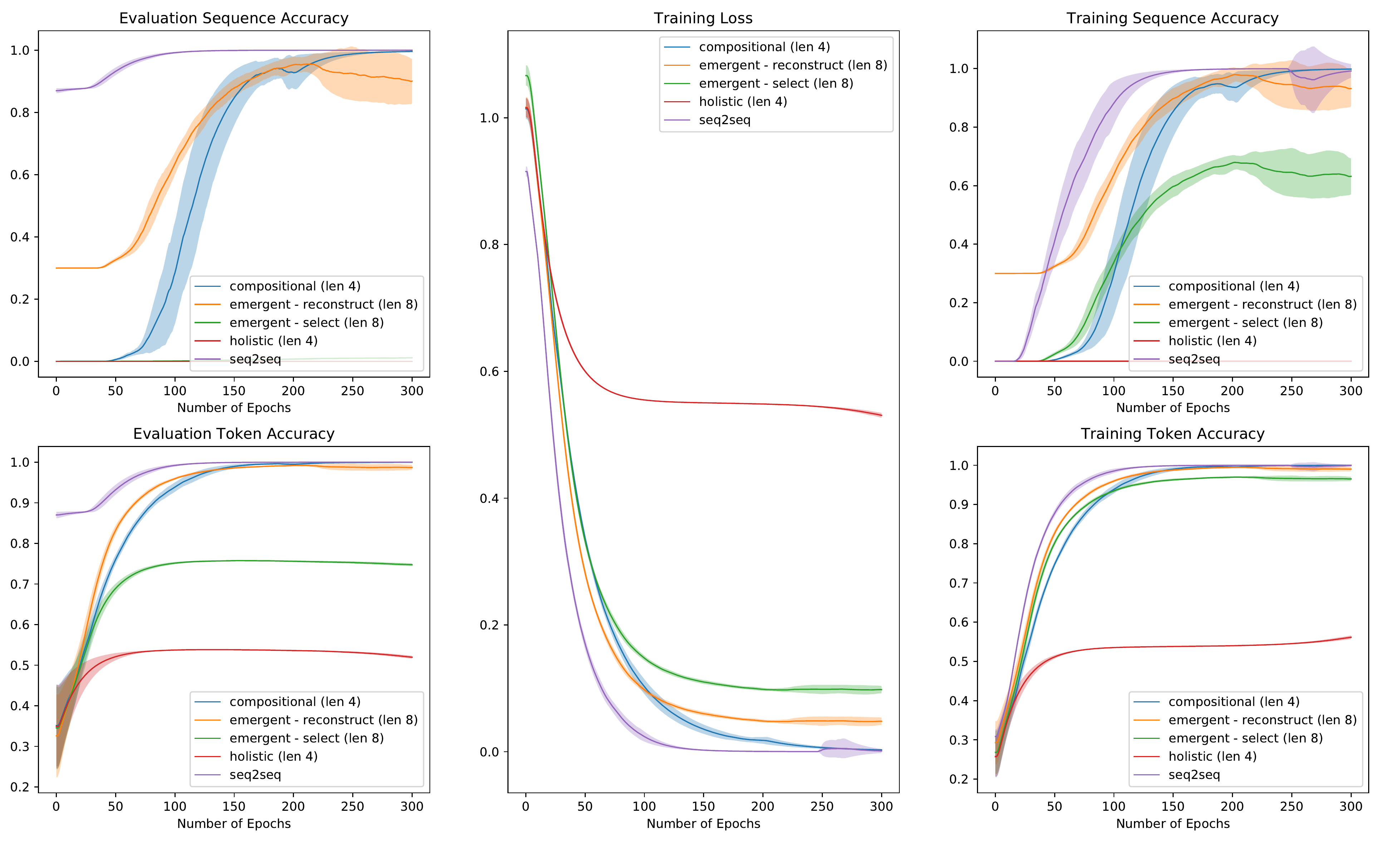}
    \caption{Learning and performance curves of different languages for listeners in Set-Reconstruct game.}
    \label{fig4.0:listener_learning_generalise_gen}
\end{figure}

\begin{figure}[!h]
    \centering
    \includegraphics[width=\textwidth]{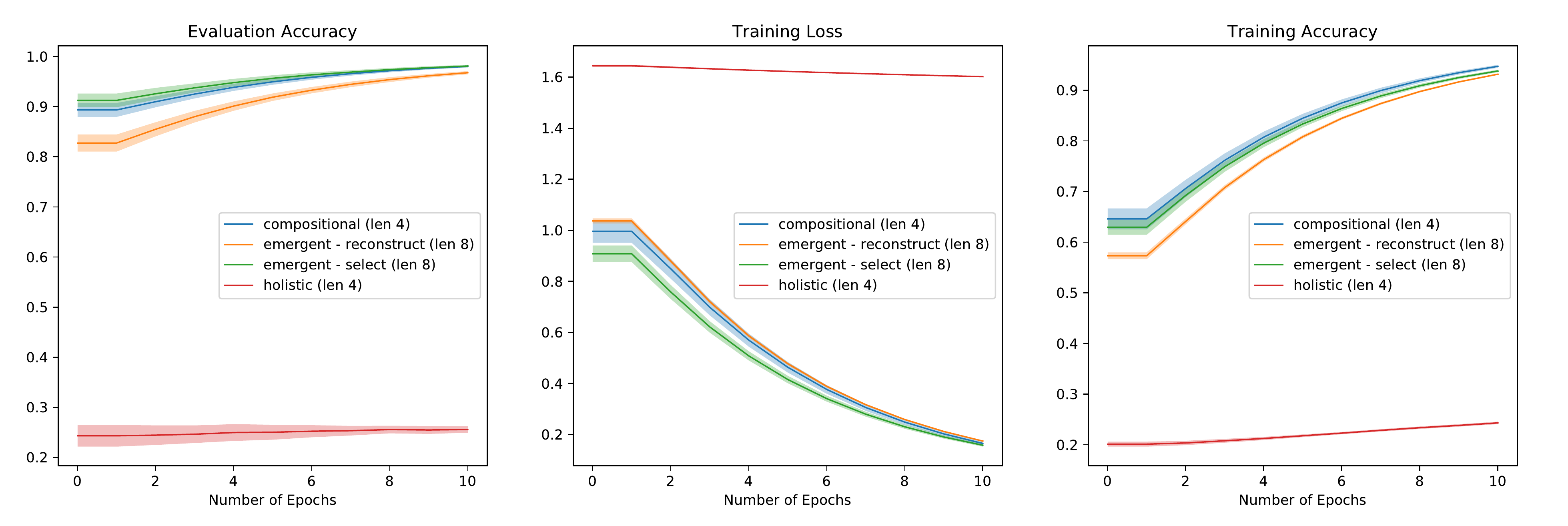}
    \caption{Learning and performance curves of different languages for listeners in Set-Select game.}
    \label{fig4.00:listener_learning_generalise_ref}
\end{figure}

Therefore, we train many randomly initialised listeners with several kinds of languages: i) compositional language; ii) an emergent language invented by other agents; iii) holistic language. The game settings are $|M|=8$, $|V|=10$, $|\mathcal{O}|=4$, $|N_{o}|=10$. We expose listeners with only the training set (which contains $80\%$ sets of objects randomly sampled from the whole meaning space), but test their performance on evaluation set (which contains unseen $20\%$ sets of objects left in the meaning space). Then, the generalisation ability of languages can be measured as the performance on evaluation set.Learning and performance curves of different languages on Set-Reconstruct and Set-Select games are given in Figure \ref{fig4.0:listener_learning_generalise_gen} and Figure \ref{fig4.00:listener_learning_generalise_ref} respectively.

Note that we run all experiments with 10 different random seeds to avoid the effect brought by different initialisations. The mean of 10 different runs are given as lines in Figure \ref{fig4.0:listener_learning_generalise_gen} and Figure \ref{fig4.00:listener_learning_generalise_ref}, and the shadow areas are corresponding standard deviations. Meanings of y-axes are given as the title for each sub-plot, and the numbers in parentheses are length of messages in each language. Peculiarly, ``emergent - reconstruct'' and ``emergent - select'' represent emergent languages from Set-Reconstruct and Set-Select games respectively. This is also the case for all the following figures.

It is quite surprising that, although we cannot see any significant pattern in the emergent language, it actually can be generalised to unseen sets of objects by listeners, as illustrated by the performance of listeners on evaluation dataset. Meanwhile, listeners trained with emergent languages converge faster on evaluation performance as well as training loss, although length of emergent messages ($|M|=8$) are longer than that of compositional language ($|M|=4$).

From the evaluation accuracy in Figure \ref{fig4.00:listener_learning_generalise_ref}, we could see that the emergent language in Set-Reconstruct game can be well generalised to unseen samples by listeners in Set-Select game. On the other hand, listeners in Set-Reconstruct game, however, cannot generalise emergent language from Set-Select game, which is illustrated by the evaluation accuracies in Figure \ref{fig4.0:listener_learning_generalise_gen}. This phenomenon demonstrates that the information encoded by speakers in Set-Reconstruct games are richer than the information encoded by speakers in Set-Select games. As the only difference between candidates in Set-Select game is the numbers of different types of objects, if we assume that the different numbers of objects are encoded by speakers in Set-Select game, then we can infer that speakers in Set-Reconstruct game would encode more than only numeric concepts. Another possible explanation is that speakers in Set-Reconstruct encode only numeric concepts, then speakers in Set-Select game would encode less than that. However, as agents could always get almost perfect performance on both training set and evaluation set, it is reasonable to believe that the later assumption is less possible.

\subsection{Section Conclusion}
\label{ssec4.2.5:sec_conclusion}

To sum up from all above, although we cannot find observable patterns in emergent languages under various game settings, the emergent languages are actually easier for agents to learn and also can be generalised to unseen sets of objects. Thus, based on the previous topological similarity metrics and significance test, we claim that the emergent languages do capture the underlying structure of meaning space.


\section{Learning Speed \& Iterated Learning}
\label{sec4.3:learning_speed}

From the previous sections, we could see that the emergent languages can reflect the underlying structure of meaning spaces, although they may not be as compositional as our natural languages. Thus, we are further curious about the motivation of the emergent language. Or, to say, the reasons why computational agents prefer to communicate in such a ``non-natural''\footnote{From a human perspective, it is not like how we communicate numeric concepts through natural language.} way.

\subsection{For Listener}
\label{ssec4.3.1:learning_listener}

\begin{figure}[!h]
    \centering
    \includegraphics[width=\textwidth]{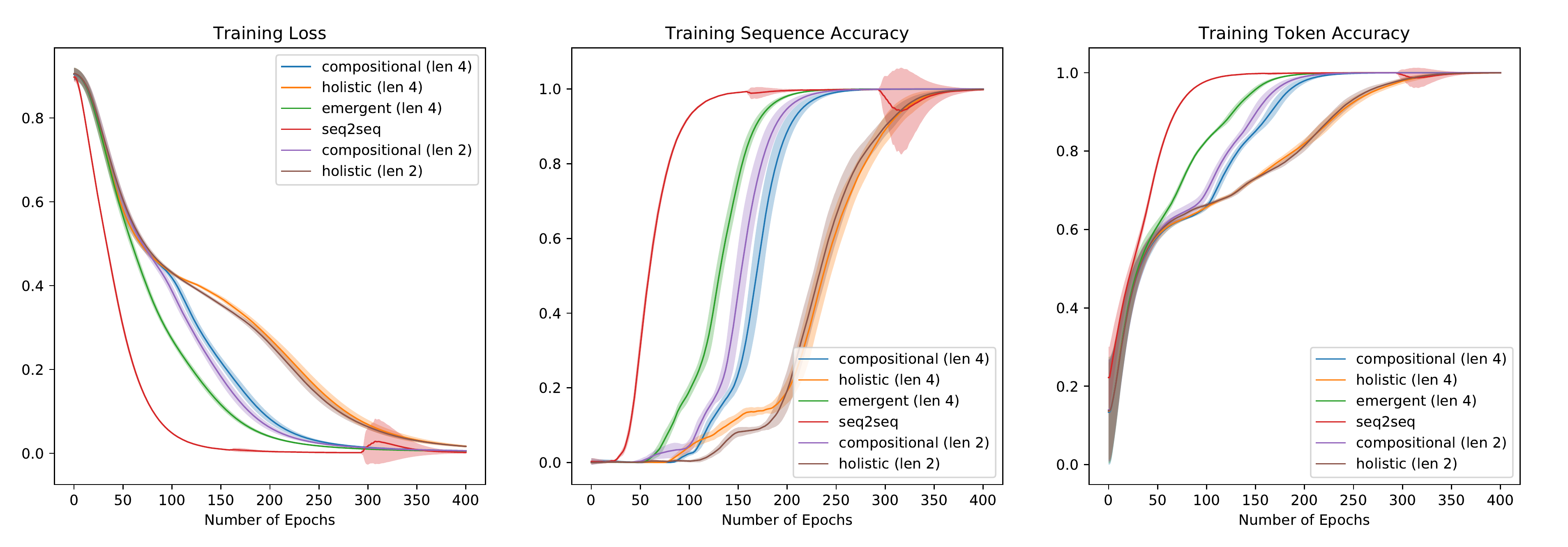}
    \caption{Learning and performance curves of different languages for listeners in Set-Reconstruct game. Numbers in the parentheses are length of messages in a language.}
    \label{fig4.1:listener_learning_generation}
\end{figure}

\begin{figure}[!h]
    \centering
    \includegraphics[width=\textwidth]{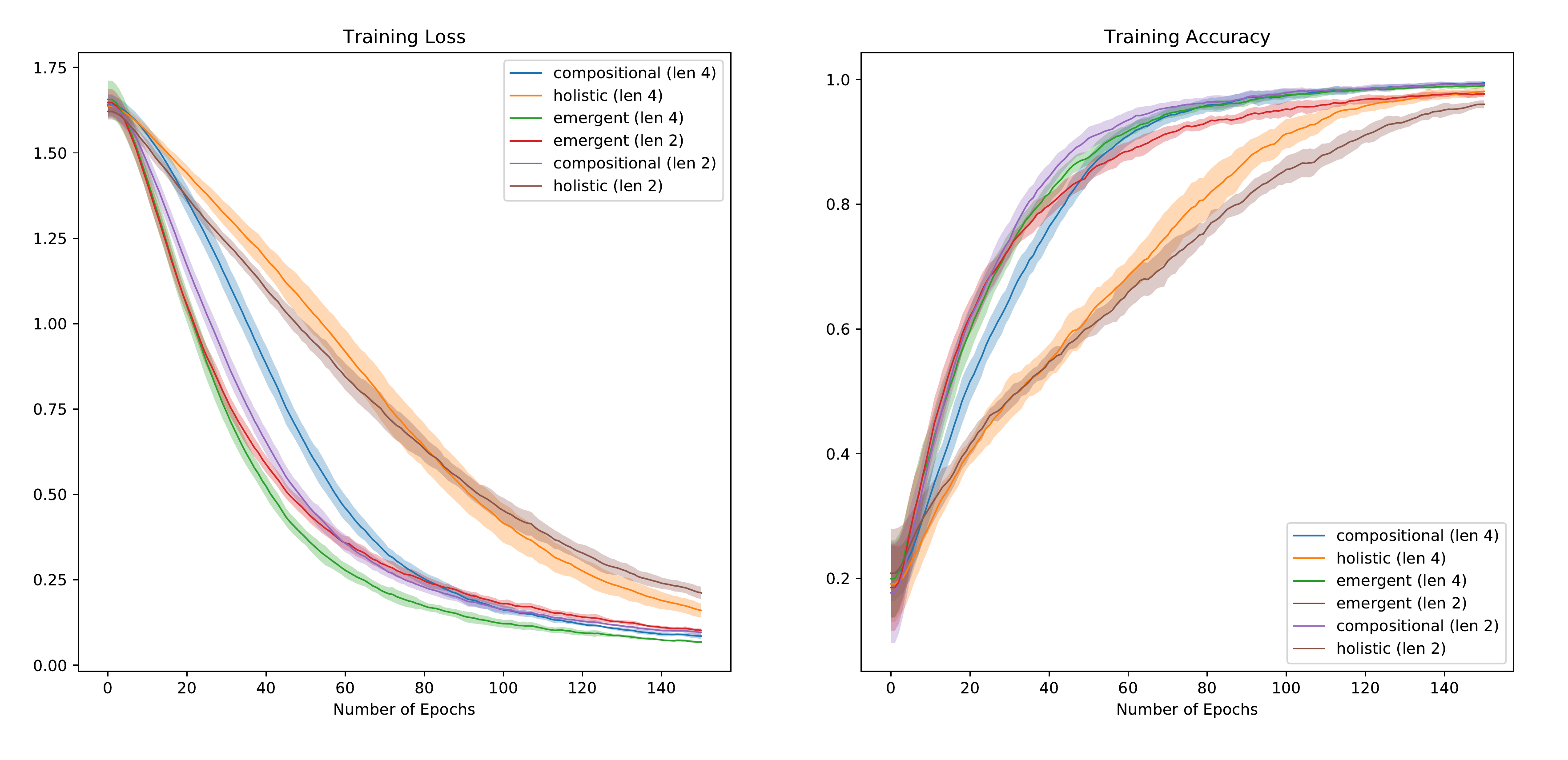}
    \caption{Learning and performance curves of different languages for listeners in Set-Select game. Numbers in the parentheses are length of messages in a language.}
    \label{fig4.2:listener_learning_refer}
\end{figure}

First we attempted to verify whether the emergent languages are the most easy ones for listeners to understand. To verify this, we first trained agents with game setting $|V|=10$, $|\mathcal{O}|=2$, $|N_{o}|=5$ and 2 different $|M|$ (2 or 4), to got 2 different emergent languages. Then, we test the learning speeds of all kinds of languages with randomly initialised new listeners in both Set-Reconstruct game and Set-Select game. Again, we run experiments with 10 different random seeds, and draw the mean and corresponding standard deviations of performance results from Set-Reconstruct and Set-Select in Figure \ref{fig4.1:listener_learning_generation} and Figure \ref{fig4.2:listener_learning_refer} respectively.

 We also trained a vanilla Seq-to-Seq model to get an upper bound of learning speeds and performance, as we assume that the original meaning space is the optimal message space for both speakers and listeners. However, as time is limited, we have not done the same experiment on Set-Select game.

From the above figures, we could easily see that emergent languages are learnt faster than compositional and holistic language in Set-Reconstruct game, while they have very similar performance to the compositional language with length 2 in Set-Select game but still have lower loss.

\subsection{For Speaker}
\label{ssec4.3.2:learning_speaker}

\begin{figure}[!h]
    \centering
    \includegraphics[width=\textwidth]{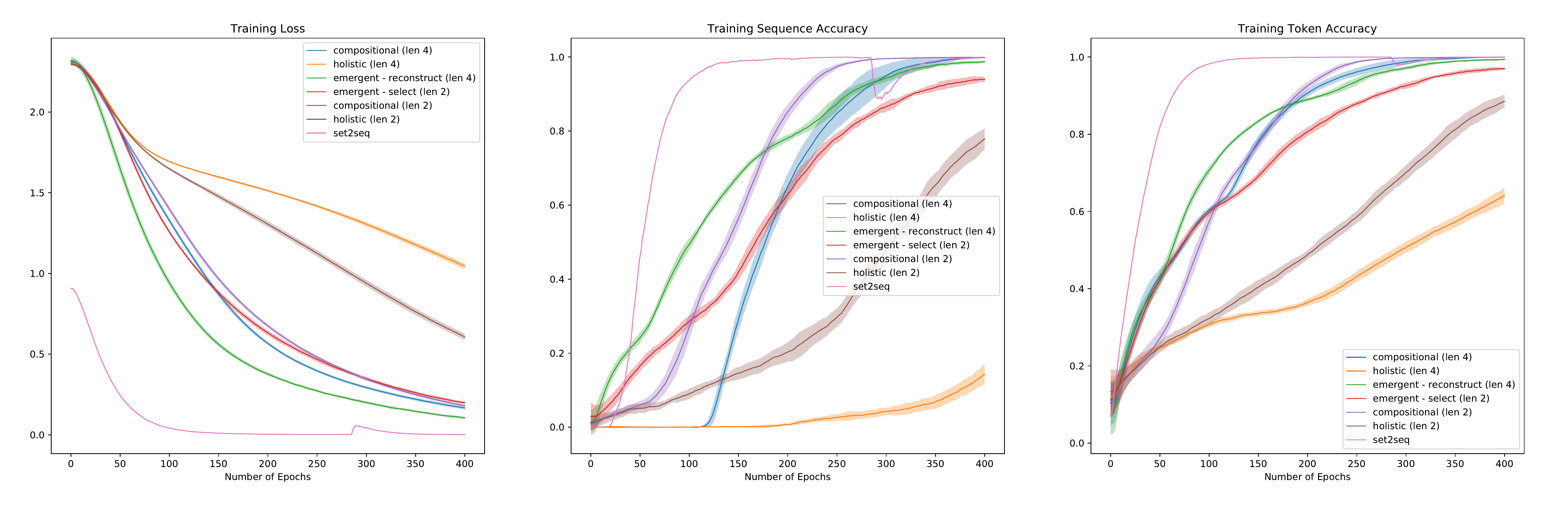}
    \caption{Learning curves of different languages for speakers in both games. Numbers in the parentheses are length of messages in a language.}
    \label{fig4.3:speaker_learning}
\end{figure}

We then test the learning speed of different languages on speaker, i.e. we randomly initialise several new speakers and let it learn to produce messages of input sets under different languages. Note that the architecture of speakers are identical in Set2Seq2Seq and Set2Seq2Choice model, and all the curves are drawn in one figure, i.e. Figure \ref{fig4.3:speaker_learning} given as follow. The game settings are the same as they are during testing listener leaning speed. One thing we need to mention is that we also test an emergent language in Set-Select game with $|M|=2, |V|=10, |\mathcal{O}|=2, |N_o|=5$ which is denoted as ``emergent (len 2)''.

Meanwhile, by comparing learning emergent languages with length $2$ and $4$, we could easily see that larger message spaces are always easier to learn than the smaller ones for emergent languages, which is counterintuitive. Considering that facts that the original meaning space is always the easiest for both speakers and listeners to learn, which is demonstrated by Figure \ref{fig4.1:listener_learning_generation}, Figure \ref{fig4.2:listener_learning_refer} and Figure \ref{fig4.3:speaker_learning}, our explanation about this phenomenon is that the larger message spaces are easier to be shaped like the original set space for agents. To be specific, as the maximum size of sets is  $|N_o|\times |\mathcal{O}|$ and the vacancies in the sets can all be represented by some special symbols (e.g. blank space in English), a larger $|M|$ would make it easier for agents to create a message space that is highly similar to the original set space and becomes easier to learn for agents.

From the above figures, it is quite clear that compositional languages always converge faster than the same sized emergent languages, which is contradictory to the situation on listener side. However, it is still the case that the emergent languages have lower losses than the same-sized compositional languages. Our hypothesis is that compositional language is a smoother function for speaker to learn and thus it is easier to be optimised. As time is limited, this phenomenon is not further discussed in this work but will be explored in the future works.

\subsection{Improvement by Iterated Learning}
\label{ssec4.3.:iterated_learning_improve}

Iterated learning framework \cite{kirby2002emergence} has been proposed to explain the emergence of language structures more than one decade. Thus, we are curious about whether it could improve the compositionality of the emergent languages in our system. However, there are several obstacles for directly applying iterated learning into our neural agents:

\begin{enumerate}
    \item we cannot feed prior probability that favours high compositional languages to neural networks;
    \item the pre-training procedure in learning phase of original iterated learning need to be re-designed, as speakers and listeners in our game are not inverse functions to each other.
\end{enumerate}

\begin{figure}[!h]
    \centering
    \includegraphics[width=\textwidth]{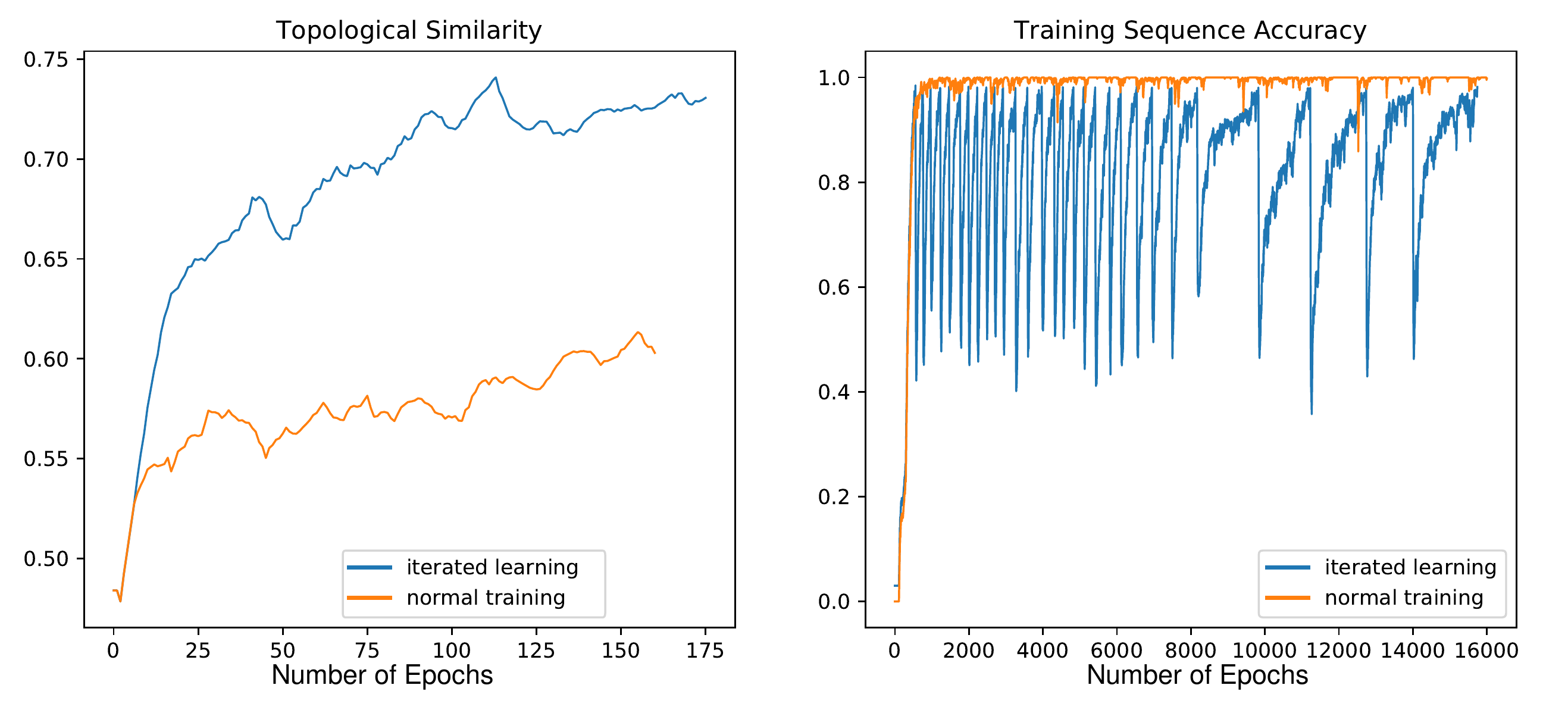}
    \caption{Topological similarity curves of iterated learning and normal training in Set-Reconstruct game. Note that one generation contains many epochs and we only test the topological similarity every 10 epochs or at the end of game playing phase, so the two plots have different scales on the x-axis. As we use early-stopping during game playing phase, every generation may contain different numbers of epochs.}
    \label{fig4.4:il_improve}
\end{figure}

Thus, we adapt iterated learning into our project, which is illustrated in Subsection \ref{ssec3.2.4:iterated_learning}, and train agent population with respect to normal training mechanism and iterated learning. The results are shown in Figure \ref{fig4.4:il_improve}. It needs to be pointed out that the distance metric for meaning space is Euclidean distance for topological similarity, and metric for message space is edit distance. Reminds that the topological similarity of compositional language under this metric is 0.38.

By comparing the curves of iterated learning and normal training, we can see a significant improvement of topological similarity in iterated learning, about $0.1$. However, although the messages emerged in iterate learning becomes more correlated with Euclidean distances between meanings, the numeric concepts in them are still not represented like numerals in natural languages.


\section{Effects of Different Representations}
\label{sec4.4:represent_effect}

Compared our results in Section \ref{sec4.1:emergence} to \ref{sec4.3:learning_speed} with previous works in grounded language learning, we argue that the different characteristics of emergent languages in our works are due to the feature representations of meanings.

To be specific, in our games, listeners need to generate object sequences or select the correct object sequence according to features representing each kind of objects. For example, the feature representation of set $\{A, B, A\}$ would be a sequence $\{[1 0], [0 1], [1 0]\}$ (assume that $|\mathcal{O}|=2, |N_o|=8$), and the corresponding message would be $\{2, 1\}=\{[0 0 1 0 0 0 0 0 0], [0 1 0 0 0 0 0 0 0]\}$ (assume that $|M|=|\mathcal{O}|=2, |V|=|N_o|=8$). Thus, to understand the message, the listener needs to correctly count the numbers of each kind of objects in the set and ground symbols to the counting results. During this procedure, there are 2 gaps between meanings (or perceptions) and messages: i) from meaning to numeric concepts; ii) from numeric concepts to messages.

To verify which step imports bias towards emergent language, we slightly change the representation of sets in Set-Select game, i.e. we directly encode the numbers of each kind of objects as one-hot vectors and concatenate them to be the representation of the whole set. Take set $\{A, B, A\}$ as example, its representation would be \textbf{vector} $[0 0 1 0 0 0 0 0 0; 0 1 0 0 0 0 0 0 0]$, whereas its message is still the \textbf{sequence} \\ $\{[0 0 1 0 0 0 0 0 0], [0 1 0 0 0 0 0 0 0]\}$. Then, it is straightforward that mapping from messages to meanings is a linear transformation and thus it should be easy for neural networks to fit.

\begin{figure}[!h]
    \centering
    \subfigure[Listener learning speed]{
        \includegraphics[width=0.48\textwidth]{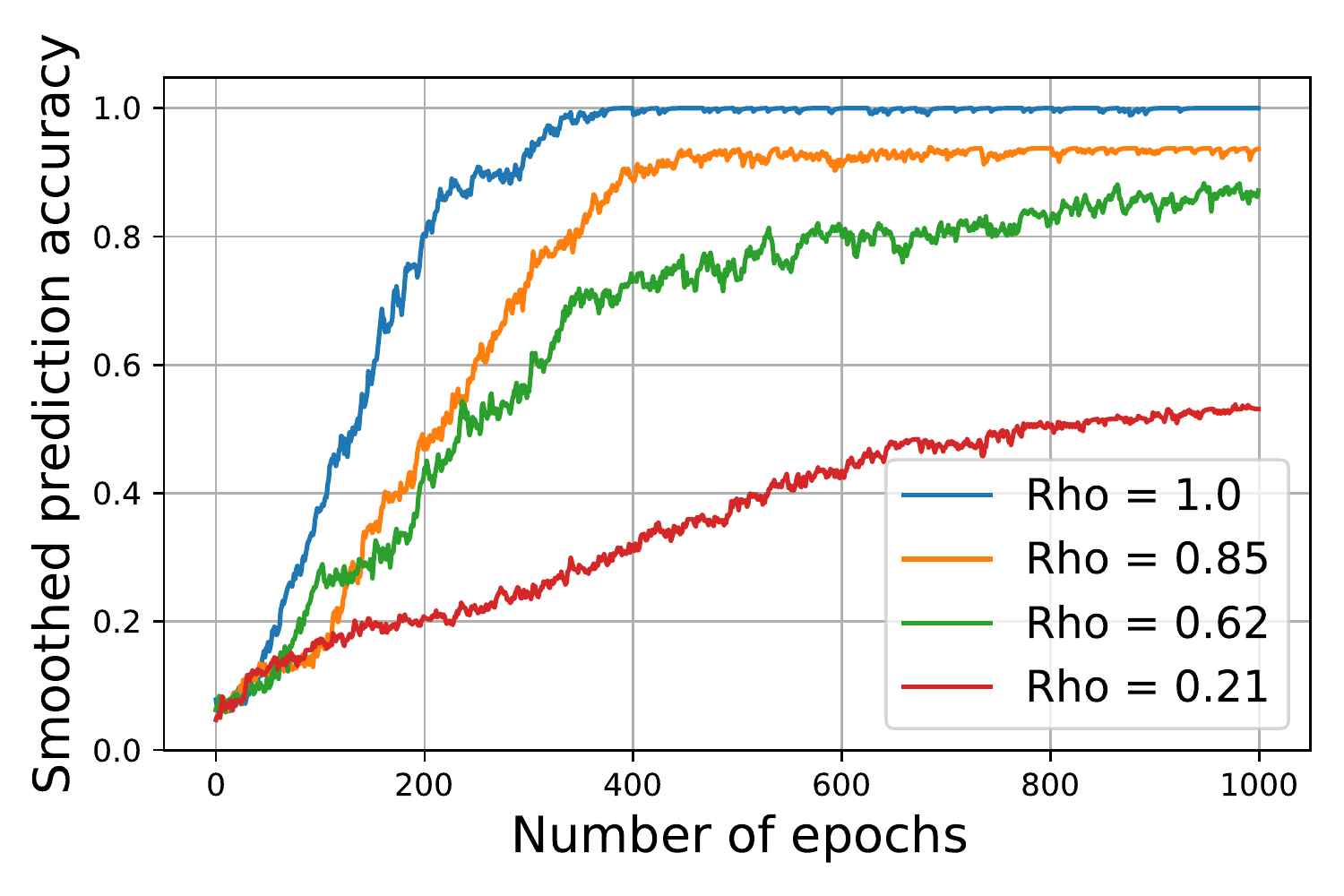}
    }
    \subfigure[Speaker learning speed]{
        \includegraphics[width=0.48\textwidth]{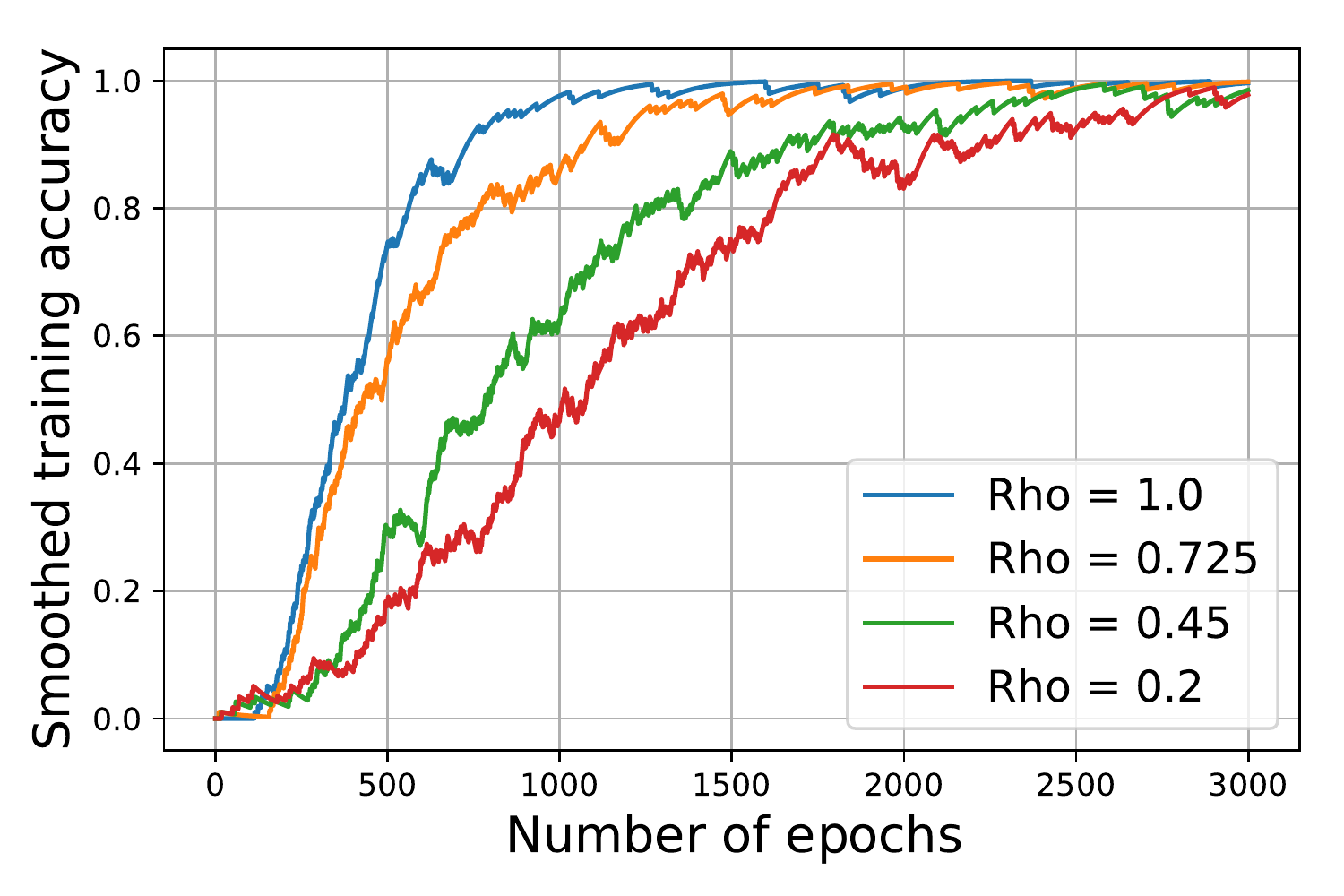}
    }
    \caption{Learning speed of languages with different compositionalities with linear feature representations.}
    \label{fig4.5:learning_speed_joshua}
\end{figure}

First of all, we test the learning speed of manually designed languages with different topological similarity scores on both speaker and listener side, and the results are shown in Figure \ref{fig4.5:learning_speed_joshua}. Note that the metric for meaning distance is Hamming distance and thus languages with higher $\rho$-values would ``look'' more like our natural language. A higher $\rho$ means that the language is more compositional from perspective of human beings, e.g. $\rho=1$ means that the language is perfectly compositional from our view.

As we can see in Figure \ref{fig4.5:learning_speed_joshua}, language with higher $\rho$-values are much more easier to learn for both speaker and listener, under the current scenario.

Then, we are curious about probability distributions of languages generation by generation. Recall that a language $\ell$ is a mapping function from meaning space $\mathcal{S}$ to message space $\mathcal{M}$, i.e. $L\in \mathcal{S} \times \mathcal{M}$. Then, we can define the probability of a language $p(L)$ as the product of probabilities of the message corresponding a given set in it, i.e. $p(\ell)=\prod_{i} p(m_i|s_i), \forall s_i\in \mathcal{S}$ where $m_i$ is the corresponding message in language $\ell$ for set $s_i$. The probability of a message given an input set is then the product of probabilities of each symbol, i.e. $p(m_i) = \prod_k p(t_{i_k}|h_s^s, t_{i_{-k}})$ where $h_s^s$ is the hidden representation of set $s_i$ and $t_{i_{-k}}$ is the symbols that appear before $t_{i_k}$.

\begin{figure}[!h]
    \centering
    \includegraphics[width=0.6\textwidth]{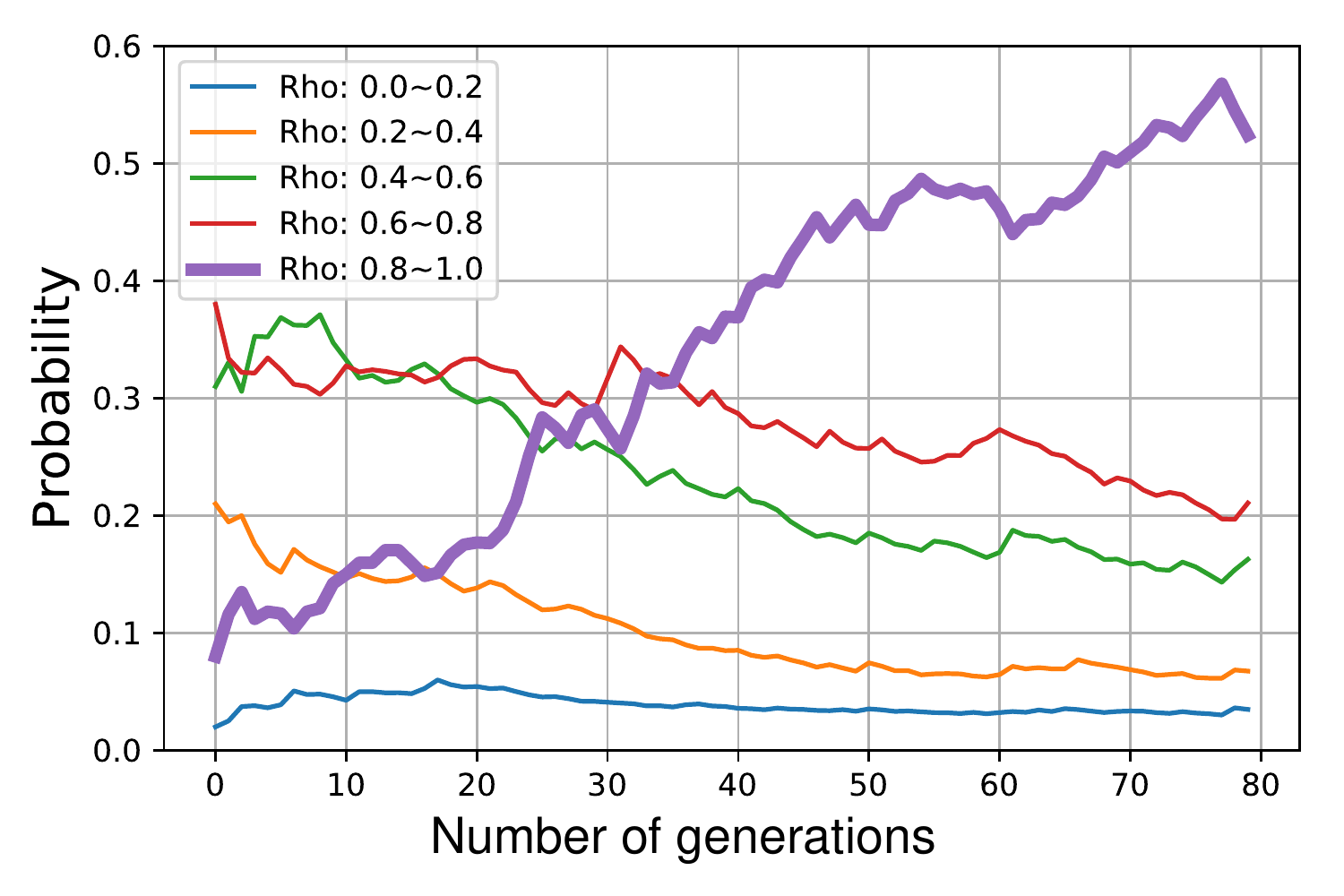}
    \caption{Changes of probabilities of languages with different $\rho$-values during iterated learning.}
    \label{fig4.6:lan_prob_IL}
\end{figure}

We track the probabilities of languages with different $\rho$-values during the iterated leaning procedure. To be specific, we manually designed many languages with different $\rho$ and calculate their probabilities at the end of each generation. The results are shown in Figure \ref{fig4.6:lan_prob_IL}.

From the above figure, it is straightforward to see that high compositional languages gradually dominate among all kinds of languages generation by generation.

To have an intuitive feeling about the final emergent language with iterated learning and current feature representations, we illustrate it in Table \ref{tab4.6:emregent_language_referential_perfect}.

\begin{table}[!h]
    \centering
    \begin{tabular}{|c|c|c|c|c|c|c|c|c|c|}
        \hline
           & 0A & 1A & 2A & 3A & 4A & 5A & 6A & 7A & 8A \\ \hline
        0B &    & yq & uq & xq & xq & zq & vq & tq & qq \\ \hline
        1B & wy & yy & uy & sy & xy & zy & vy & ty & qy \\ \hline
        2B & ws & ys & us & ss & xs & zs & vs & ts & qs \\ \hline
        3B & wt & yt & ut & st & xt & zt & vt & tt & qt \\ \hline
        4B & wu & yu & uu & su & xu & zu & vu & tu & qu \\ \hline
        5B & wv & yv & uv & sv & xv & zv & vv & tv & qv \\ \hline
        6B & wz & yz & uz & sv & xz & zz & vz & tz & qz \\ \hline
        7B & ww & yw & uw & sw & xw & zw & vw & tw & qw \\ \hline
        8B & xx & yx & ux & sx & xx & zx & vx & tx & qx \\ \hline
        \end{tabular}
    \caption{Final emergent language in linear feature representation and iterated learning.}
    \label{tab4.6:emregent_language_referential_perfect}
\end{table}

Then, it is straightforward to decipher the emergent language shown in Table \ref{tab4.6:emregent_language_referential_perfect}. Basically, the symbols appeared in the first digit represent the numbers of ``A'' and the symbols appeared in second digit represent the numbers of ``B''. Of course, the language is still not perfect compositional, as there are some repetitive messages for different meanings, such as ``3A5B'' (sv) and ``3A6B'' (sv). Besides, it worth mentioning that the same symbol still represents different meanings if it appears at different positions.

Overall, we could say the the obstacle for the emergence of compositional languages in our Set-Reconstruct and Set-Select games is that symbols in messages do not directly correspond to numeric features in the original meaning spaces. As long as the features we want the emergent language to represent is established, the agents could invent almost perfect compositional language by iterated learning.

\section{Further Discussion}
\label{sec4.5:discuss}

Comparing the experimental results in this chapter with previous work in grounded language learning, e.g. \cite{kottur2017natural, hermann2017grounded, havrylov2017emergence, mordatch2018emergence}, we propose an alternative hypothesis to explain the emergence of compositional language (some previous works call it natural language) during the autonomous communication among agents population.

First of all, we argue that the feature vectors of input experience and perceptions should be inherently disentangled, i.e. the feature vectors of these inputs should satisfy mutual exclusivity and orthogonality defined in Subsection \ref{ssec4.2.1:emergent_languages}, in order to facilitate the emergence of compositional language. Then, it could be an optimal method to use a single symbol as a feature representation of a disentangled element in feature vectors. By comparing the emergent languages in Section \ref{sec4.1:emergence} to \ref{sec4.3:learning_speed} with that in Section \ref{sec4.4:represent_effect}, it is straightforward to see that linear transformed feature representations would be much more optimal for the emergence of highly compositional languages. However, as lots of previous use images as the perceptions for speakers, there is still a gap between our 2 representing methods. Without further experiment, we are not sure about whether the emergence of compositional languages of those works are caused by that convolutional neural networks (CNN) can spontaneously encode images into disentangled representations. Previously, it has been widely believed that the success of unsupervised learning for CNNs depends on that models can automatically establish disentangled representations \cite{bengio2013representation}. However, recently, this common assumption become questioned and challenged by researchers \cite{locatello2018challenging}. Thus, our hypothesis is that the emergence of compositionality is actually highly related to the disentangled representation of models.

Secondly, we argue that iterated learning is an effective method to amplify inductive bias into multi-agent autonomous communication systems, and thus improve the compositionalities of emergent languages. Considering the discoveries in \cite{locatello2018challenging}, we claim that the compositional languages are highly correlated with the appearance of disentangled representations. Further, inductive bias towards compositionalities of different kinds of symbols (which correspond to words in natural languages) should be introduced to different spaces. For example, inductive bias towards the compositionality of symbols corresponding to objects/attributes that physically exists in real/virtual world can be introduced by iterated learning, as the feature values of these objects/attributes are mutually exclusive and independent (or to say, they are inherently disentangled). On the other hand, compositionality of function words, such as numerals in our project, requires the agents to first encode the input features in some specific ways and obtain disentangled representations. Thus, without specially designed training mechanism or data samples that could introduce such pressure, it is natural for agents to invent effective but non-natural ``languages'' during their autonomous communication. 

%% file: chapter5.tex
\chapter{Conclusions}
\label{ch5:conclusion}

\section{Express Numeric Concepts}
\label{sec5.1:numeric_represent}

With all experimental results shown in Chapter \ref{ch4:results_analysis}, we could conclude that the models illustrated in Chapter \ref{ch3:game_model} can successfully transmit numeric concepts in whichever Set-Reconstruct or Set-Select game proposed in Chapter \ref{ch3:game_model}. Although the emergent languages are not compositional from the perspective of humans, they do capture the underlying structure of meaning space and reflect it into messages consist of sequences of discrete symbols, which is measured by the Euclidean distances between meaning pairs. Furthermore, the messages expressing same numeric concepts have higher similarities to each other, which is measured the BLEU score defined in Section \ref{sec3.3:metrics}. More importantly, the emergent languages can be successfully generalised to unseen meanings and they are not only effective but also efficient, as models can fit to them faster than other languages.

Therefore, we claim that the agents capture the numeric concepts during cooperating to complete the games , and successfully transmit these numeric concepts with a non-natural language.

\section{Role of Iterated Learning}
\label{sec5.2:iterated_learning}

By transforming iterated learning to train our DL-based agents, it successfully improves the compositionality of emergent languages, which is measured by Euclidean distances in meaning space and BLEU score in message space, in our original set representations of objects. Then, by taking vectors that directly encode quantities of different kinds of objects as the input for speakers, the emergent languages become almost perfectly compositional under iterated learning.

Therefore, we claim that iterated learning is an effective method to improve the compositionality of emergent languages, w/o inherently disentangled feature representation of inputs. Even thought the emergent compositionality may not correspond to what it is in human natural languages.

\section{Future Works}
\label{sec5.2:future_work}

With the current exploration, there are still several open questions in our work and thus several interesting and meaningful future works:

\begin{enumerate}
    \item \textbf{Generalisation and meta-learning}: \cite{smith2013linguistic} claims that language structure is an evolutionary trade-off between simplicity and expressivity. We assume that generalisation is another form of this trade-off. Further, emergence of numerals is a good candidate for discovering the role of generalising pressure in language evolution, as numerals can be used for whatever kind of objects. More importantly, such pressure can be formalised by meta-learning.
    \item \textbf{Feature representations}: As discussed in Section \ref{sec4.4:represent_effect}, different kinds of representations have a strong effect on the compositional form of emergent languages. Argued by \cite{locatello2018challenging}, representation learnt without supervision are not disentangled. We further assume that inherently disentangled elements are not only important in the input feature space but also in the parameter feature space. Or, to say, some words in our natural languages directly correspond to elements in input feature representations, while others may correspond the features of specific functions, e.g. agents need to learn counting (a function) in our games.
\end{enumerate}